\documentclass[sigconf]{acmart}

\usepackage{tcolorbox, soul}
\usepackage{caption}
\usepackage{subcaption}
\newtheorem{definition}{Definition}

\newtheorem{conjecture}{Conjecture}
\newtheorem{query}{\textbf{Query}}
\newtheorem{response}{\textbf{Response}}

\newcommand{\ourapproach}{\textit{Covertex}}

\AtBeginDocument{%
  \providecommand\BibTeX{{%
    \normalfont B\kern-0.5em{\scshape i\kern-0.25em b}\kern-0.8em\TeX}}}

\setcopyright{acmlicensed}
\copyrightyear{2024}
\acmYear{2024}
\acmDOI{XXXXXXX.XXXXXXX}

\renewcommand\footnotetextcopyrightpermission[1]{} % removes footnote with conference information in first column

\begin{document}

\title{Co(ve)rtex: ML Models as storage channels and their (mis-)applications}


\author{Md Abdullah Al Mamun}
\affiliation{%
  \institution{UC Riverside}
  \city{Riverside, Califronia}
  \country{USA}}
\email{mmamu003@ucr.edu}

\author{Quazi Mishkatul Alam}
\affiliation{%
  \institution{UC Riverside}
  \city{Riverside, Califronia}
  \country{USA}}
\email{qalam001@ucr.edu}

\author{Erfan Shayegani}
\affiliation{%
  \institution{UC Riverside}
  \city{Riverside, Califronia}
  \country{USA}}
\email{sshay004@ucr.edu}

\author{Pedram Zaree}
\affiliation{%
  \institution{UC Riverside}
  \city{Riverside, Califronia}
  \country{USA}}
\email{pzare003@ucr.edu}

\author{Ihsen Alouani}
\affiliation{%
  \institution{CSIT, Queen's University Belfast}
  \city{Belfast}
  \country{UK}}
\email{i.alouani@qub.ac.uk}

\author{Nael Abu-Ghazaleh}
\affiliation{%
  \institution{UC Riverside}
  \city{Riverside, Califronia}
  \country{USA}}
\email{nael@cs.ucr.edu}

\newcommand{\nael}[1]{\textcolor{blue}{ \@ Nael: #1}}
\newcommand{\ihsen}[1]{\textcolor{cyan}{ \@ Ihsen: #1}}
\newcommand{\mamun}[1]{\textcolor{violet}{ \@ Mamun: #1}}
\newcommand{\mishkat}[1]{\textcolor{green}{ \@ Mishkat: #1}}
\newcommand{\erfan}[1]{\textcolor{orange}{ \@ Erfan: #1}}

\renewcommand{\shortauthors}{Mamun, Alam, Shayegani, Zaree, Alouani and Abu-Ghazaleh et al.}


\begin{abstract}
Machine learning (ML) models are overparameterized to support generality and avoid overfitting. The state of these parameters is essentially a "dont-care" with respect to the primary model provided that this state does not interfere with the primary model. In both hardware and software systems, don't-care states and undefined behavior have been shown to be sources of significant vulnerabilities.  In this paper, we propose a new information theoretic perspective of the problem; we consider the ML model as a storage channel with a capacity that increases with overparameterization.  Specifically, we consider a sender that embeds arbitrary information in the model at training time, which can be extracted by a receiver with a black-box access to the deployed model. We derive an upper bound on the capacity of the channel based on the number of available unused parameters. We then explore black-box write and read primitives that allow the attacker to: \textbf{(i)} store data in an optimized way within the model by augmenting the training data at the transmitter side, and \textbf{(ii)} to read it by querying the model after it is deployed. We also consider a new version of the problem which takes information storage covertness into account. Specifically, to obtain storage covertness, we introduce a new constraint such that the data augmentation used for the write primitives minimizes the distribution shift with the initial (baseline task) distribution. This constraint introduces a level of "interference" with the initial task, thereby limiting the channel's effective capacity. Therefore, we develop optimizations to improve the capacity in this case, including a novel ML-specific substitution based error correction protocol. We analyze the achievable capacity for different size networks and models, demonstrating significant capacity to transfer data with low error rates. Our work offers new tools to better understand and mitigate  vulnerabilities of ML models, especially in the context of increasingly large models. 
\end{abstract}



%\ccsdesc[500]{Security and privacy~Software and application security}

%\keywords{Neural channel capacity, Covert channel, Large language model, Error correcting codes, Pruning, Fine tuning}

\maketitle

\section{Introduction}\label{sec:intro}

Machine learning (ML) models deliver state-of-the-art performance across many application areas including computer vision \cite{simonyan2014deep,redmon2016yolo9000,he2016deep,alom2018recurrent}, natural language processing (NLP) \cite{deng2018deep,chowdhary2020natural,hirschberg2015advances}, robotics \cite{pierson2017deep,garcia2007evolution,mnih2013playing}, autonomous driving \cite{al2017deep,teichmann2018multinet,bojarski2016end}, and healthcare \cite{miotto2018deep,sahlsten2019deep,arcadu2019deep}.   With their increasing deployment for critical applications, a number of threat models have been identified that can affect the security of the model or the privacy of the data that is used to train it.  For example, adversarial attacks~\cite{carlini2017towards,huang2011adversarial,goodfellow2014explaining,moosavi2016deepfool} and poisoning attacks~\cite{munoz2017towards,sun2019can,bagdasaryan2020backdoor,biggio2012poisoning,fung2018mitigating,naseri2020toward} compromise security by causing the model to misclassify to the attacker's advantage.  Similarly, privacy related attacks can leak private information about the data used in training the model~\cite{shokri2017membership,nasr2019comprehensive,zhang2020gan,hitaj2017deep,melis2019exploiting,luo2021feature}.  

%Since the development of deep learning, new generation architectures are consistently increasing in terms of size, i.e., architecture depth and number of parameters. In fact,  often models are trained using architectures that contain more parameters than the training samples of the dataset \textcolor{blue}{some examples here/ maybe a trend figure? }. Such over-parametrized models have been extensively studied in recent years, and the virtues of over-parametrization have been established from both the statistical perspective, via the double-descent phenomenon, and the computational perspective via the structural properties of the optimization landscape. 

%Machine learning (ML) in general, and Deep Neural Networks (DNNs) in particular, deliver state-of-the-art performance across many application areas including computer vision \cite{simonyan2014deep,redmon2016yolo9000}, natural language processing (NLP) \cite{deng2018deep}, robotics \cite{pierson2017deep}, autonomous driving \cite{al2017deep}, and healthcare \cite{miotto2018deep}.  
 The virtues of over-parameterization in machine learning have been established from a statistical point of view as a necessary technique to deal with high-dimensional data.  New generations of ML architectures continue to emerge with increasing size over time; for example, the original stable diffusion models had less than 1 Billion parameters, while Dall-E has $12$ B parameters \cite{dalle,dalle2}. Large Language Models (LLMs) are also growing, with GPT-4 rumored to have over trillion parameters~\cite{gpt4}.  As the model sizes increase, the number of available unused parameters also continues to increase. %Besides, from a computational perspective, overparametrisation improves the numerical stability of the optimization methods used for training ML models.

%ML models are typically overparameterized to promote generalization and avoid overfitting.   Interestingly, prior works have shown that these additional parameters that are unused by the model can be used both for malicious  (for example, hiding a model covertly within a trained model~\cite{salem2021get}, or exfiltrating training data~\cite{song2017machine}), or beneficial purposes (for example, watermarking a model~\cite{darvish2019deepsigns,rouhani2018deepsigns,adi2018turning}). 

The lottery ticket hypothesis (LTH), a seminal paper in machine learning, demonstrated that for an ML model undergoing training, there exists winning tickets, i.e., smaller subnetworks which suffice on their own to capture the trained model~\cite{lottery}.  Thus, once the model is trained, many of the parameters, i.e., those are not part of the winning ticket, can be considered {\em unused} 
during inference. We identify these "spare" parameters of the initial (non-pruned) model as Unused Parameters (UPs). The conceptual implication is that the state of these parameters does not matter (or is a don't care) provided it does not interfere with the results of the winning ticket.  %We approach these problems from \textbf{a novel perspective, where we view machine learning models as communication channels} ({\`a} la Shannon). This enables us to borrow concepts and methods from communication and security theory  to study these   issues, such as the framework of covert channel attacks as well as   the concept of `don't-care states' (the extra-capacity of the communication channel). Indeed, i

In both software~\cite{wang2013towards} and hardware~\cite{fern2015hardware} systems, \textbf{undefined behavior and don't-care states} have been shown to be potential sources of vulnerabilities.  If attackers can control the state of these parameters, without affecting the baseline model, they may be able to change the state of the network to their advantage covertly.  In fact, several prior works have shown that UPs can be used both for malicious and benign purposes. For example,~\cite{salem2021get} shows that it is possible to hijack a model for a separate task. One other possible threat is to exfiltrate private training data by abusing the model capacity~\cite{song2017machine}. Other works establish that this can be used for beneficial purposes such as watermarking ~\cite{darvish2019deepsigns,rouhani2018deepsigns,adi2018turning}.  

%Given the potential vulnerabilities posed by UPs, 
Our goal in this paper is to investigate the potential misuse of ML models overparametrization.  In this threat model, the attacker changes the training to affect the state of these "don't care" parameters to their advantage.  We propose to understand and analyze the problem as a storage/communication channel. In the proposed approach, UPs can be viewed as an additional capacity beyond the baseline task, which can be abused by adversaries to store data covertly within the model, stealthily augmenting the model. We build on the previous work and explore using the spare capacity as a storage channel between an entity (sender) that trains the model and stores data in the channel, and another entity (receiver) that attempts to retrieve this data through access to the trained model; we call this channel {\em \ourapproach}.

\ourapproach~ can be used within a threat model in which a malevolent ML training as a service maliciously trains a model on behalf of a customer, but has no access to exfiltrate the private training data through direct communication.  Instead, the service stores private information in the unused parameters of the model through augmenting the training dataset.  Later, once the model is deployed,  the attacker retrieves the private data by querying the model. We first derive an upper bound on the capacity of the channel based on the number of unused (and therefore prunable) parameters. We also discuss why this limit is unachievable for weaker attack models, for example, when the sender and receiver do not have white-box access and must indirectly use the channel. In Section~\ref{blackbox}, we explore how to store values in the channel with only black-box access. Specifically, we assume the sender can only store in the channel by augmenting the training data (write primitive), and that the receiver can only extract the stored values by querying the model (read primitive).  We introduce optimizations to improve the performance of the channel.  For example,  we use \textit{Dynamic \ourapproach~(\ourapproach-D}) to differentially reduce the number of patched samples during training for storing data, consuming less capacity.

One drawback of the channel we explored so far is that the adversarial training samples used to store the data in the model are out-of-distribution and easy to identify.  Thus, we consider an alternative threat model where the attacker is limited to making the input data similar to the baseline data to avoid detection (Section~\ref{covert_channel}). Since the input sequences are covertly embedded in inputs that appear to be in-distribution, the efficiency of the channel will be lower. We develop approaches to improve the channel quality using multiple reads, as well as a novel error correction code that takes advantage of the nature of the model. Section~\ref{sec:defense} discusses potential mitigations against the storage channel and misuse of UPs.  

Our work builds on initial work by Song et al.~\cite{song2017machine} who were the first to demonstrate the transfer of private data through an ML model.  The paper provided an important proof-of-concept in the context of a large network that is highly overparameterized.  Our work systematically explores the available capacity, introducing different optimizations that together substantially increase capacity.  Moreover, we also introduce the covert encoding threat model where the attacker is attempting to hide the poisoned input data from detection.  We discuss this and other related works in Section~\ref{sec:related}.
%did not characterize the capacity or attempt to come up with efficient approaches to store the data.

The contributions of this paper are as follows:
\begin{itemize}
\setlength{\itemsep}{0in}
    \item We develop new black-box modulation techniques to store and later retrieve data in the unused parameters of the model, without affecting its primary function.  
    \item We also explore a version of the problem where data is stored covertly, making it difficult to identify anomalous input samples.  
    \item We develop optimizations to improve the the capacity and performance of both the baseline channel and the covert channel.  These include dynamic encoding and data augmentation, as well as novel error correction algorithms for the covert channel.

    \item We demonstrate high capacity and channel quality, significantly higher than prior work.  For example, in Resnet-50, trained for CIFAR-10, we are able to store 900K of random data digits (about 300KBytes) with high accuracy for the baseline model and over 25K digit capacity with good accuracy for the covert channel.

\end{itemize}

\noindent
\textbf{Disclosures and ethics.} We disclosed our findings to Guardrails AI and Nvidia. We executed all experiments within our local sessions, preventing any impact on external public users.

%------

\section{Assumptions and Threat model} \label{sec:threat}

\noindent\textbf{Attacker's objectives.} We consider an adversary that aims to leverage the "don't care" state in overparametrized models to store arbitrary information covertly within the model, thereby building a covert storage channel. The attacker wants to exploit the channel's capacity to store as much (potentially sensitive) data as possible either to bypass communication security measures (e.g. air gap), or to evade detection by defense mechanisms (e.g. Guardrails in LLMs). Importantly, exploiting the channel capacity is bounded by a constraint on the victim model's baseline accuracy; it should not affect the model performance in a significant manner. We also investigate the case where the attacker wants the channel establishment process to be covert. 

\noindent\textbf{Attacker's capabilities.} The attack has three parts with different attacker access assumptions: \textbf{(i)} Prior to training, the attacker has access to the training code/framework. They do not have access to the secret data at this point.  \textbf{(ii)} During training, the adversary has access to secret data (perhaps the training data) but is unable to exfiltrate it by direct communication (e.g., attacker is a training service without network access).  The attacker therefore uses training to store the data covertly in the model.  \textbf{(iii)} Finally, once the model is deployed,  perhaps as a
cloud based service, the attacker now has access only to query the model, and is able to recover the covertly stored data by querying the model and observing the outputs.

%The threat model in general assumes a training side attacker that is able to store information covertly in the model, without affecting its baseline performance significantly.  The attacker is then able to recover the stored data by querying the model after it is deployed.  Such a scenario can arise when a malicious training service is unable to communicate private training data directly (e.g., provisioned without network access).  In this case, the service can exfiltrate private data by covertly storing it in the trained model using our storage channel.  Once the model is deployed, perhaps as a cloud based service, the attacker is then able to recover the data having access only to query the model.  

\noindent\textbf{Attacker's knowledge.} The threat model requires that the attacker fixes an address space, which is a sequence of inputs (say patched images) that are used to index stored data. To store a value at an address, the corresponding adversarial input image is labeled with the stored data, and added to the training data. 
Once the model is deployed, the attacker queries the model with the same sequence of input images to recover the data. Note that the address input pattern is independent of the data (for example, the address pattern may be generated by a process known to both the sender and receiver); the data is the only secret being stored in the model.

%------

\section{Intuition and Problem Formulation}\label{theoritical_limit}

Consider a network trained for a baseline model, that finds a particular lottery ticket implementing the function $F_{primary}$ (e.g., classification).  An attacker that seeks to control the unused parameters to create a network to support a new augmented super function $F^*$, either by manipulating the model parameters directly (white-box access) or indirectly by adding to the training data.  Under the constraint of our current application (a storage channel), $F^*$ should approximately compute $F$ if presented with an input sample that matches the valid/active distribution for $F$. For example, for a network trained on MNIST, these samples are valid hand written digit images that we expect to be classified into one of the 10 classes even.  However, for inputs outside the distribution, which are outside the domain of $F$ (for example, some images that are not clearly from the MNIST distribution), we would like $F^*$ to serve a different purpose.  With respect to $F$, these inputs are outside the distribution and are therefore considered "don't care" and not typically included in a testing set to evaluate $F$; therefore, an attacker is able to use them to covertly store data in the network. 

Note while the mechanics of the attack that adversarially manipulate training data to change the classifier is similar to poisoning, the goals are different:  traditional poisoning changes the baseline model function $F$, for example to install a backdoor, changing the output for valid inputs in the domain of $F$.  In contrast, our proposed covert channel attempts to preserve the baseline model $F$, but augments/extends it to provide attacker controlled outputs for don't-care inputs.  
This stored data may be described as a new model $F_{covert}$: provided that the distribution of the inputs used to store the data are from a different distribution from the primary inputs of the model, and provided the model has sufficient parameters, the model is able to learn the super-function $F^{*}=F_{primary}\bigcup F_{covert}$.  When inputs from the primary distribution of the model are presented, $F_{primary}$ is computed, while the model responds with stored data (as the output of the classifier) when an input from the $F_{covert}$ distribution is used. Next, we formally describe how the available parameters are used as a covert channel.

% is considered as a communication channel
We focus on networks with the ReLU activation, $\sigma (x)= \texttt{max}\{ x, 0\}$. We define a network $h:  \mathbb{R}^d \rightarrow \mathbb{R}$ of depth $l$ and width $n$ (for simplicity we consider all layers of the network to be of the same width ) such that: $h(x)=h^{(l)} \circ \cdots \circ h^{(1)}(x)$, where: \\$h^{(1)}(x)=\sigma\left(W^{h(1)} x\right) \text { for } W^{h(1)} \in \mathbb{R}^{d \times n} \text {, and}\\ h^{(i)}(x)=\sigma\left(W^{h(i)} x\right) \text { for } W^{h(i)} \in \mathbb{R}^{n \times n} ~ \forall ~1<i<l$ \text{, and}\\ $h^{(l)}(x)=W^{h(l)} x \text { for } W^{h(l)} \in \mathbb{R}^{n \times 1}\text{.}$

    \begin{definition}\label{def:Cap} -- \textbf{Spare Subnetwork.} 
        Let $\widetilde{h}$ be a pruned subnetwork from a dense (overparametrized) model $h$ of width $n$ and depth $l$, with weights $W^{\widetilde{h}(i)}:=B^{(i)} \odot W^{h(i)}$ for some mask $B^{(i)} \in\{0,1\}^{n_{\text {in }} \times n_{\text {out }}}$.
        We define the Spare Subnetwork $\mathcal{S}$ as the complementary of $\widetilde{h}$, i.e., a model of width $n$ and depth $l$, with weights $W^{\mathcal{S}(i)}:=\Bar{B}^{(i)} \odot W^{h(i)}$. 
        
        %the number of zeros in the mask $B$. It can be expressed as:    $\mathcal{S}= HW(\Bar{B})$,

      %where $HW(.)$ is the Hamming weight operator and $\Bar{B}^{(i)}= 1$ \text{if} $B^{(i)}= 0 ~ \text{, and} \Bar{B}^{(i)}= 0 ~\text{if} {B}^{(i)}= 1 $
    \end{definition}

\noindent \textit{\textbf{Remark 1.}} The existence of Spare Subnetworks is a direct result of The Lottery Ticket Hypothesis \cite{lottery}.

\begin{definition}\label{def:mem} -- \textbf{($\ell,\delta$)-Reception.}
We say that an example $x$ can be ($\ell,\delta$)-received from a model $h_{\theta}()$ if there exists an efficient algorithm $\mathcal{R}$ (that does not have $x$ as input) such that $\hat{x}= \mathcal{R}(h_{\theta})$ has
the property that $\ell(x, \hat{x}) \leq \delta$.
\end{definition}

\noindent \textit{\textbf{Remark 2.}} Definition \ref{def:mem} is directly inspired by Definition 1 in \cite{Carlini_usenix23}. However, while Carlini et al.~\cite{Carlini_usenix23} consider extracting samples from the original training dataset of the model, our objective is more general, i.e., we consider transmitting and receiving arbitrary data through the model. From this perspective, the work in \cite{Carlini_usenix23} corresponds to the case where $X = \mathbf{S}$ in Definition \ref{def:NC}.

\begin{definition}\label{def:NC} -- \textbf{($\varepsilon, \ell,\delta$)-Neural Channel.}
Let $\varepsilon \in (0,1)$, and let $h_{\theta}(.)$ be an overparametrized neural network with parameters $\theta$ %trained for a task $\mathcal{T}$ defined by 
efficiently trained on a data distribution $(X, Y) \sim \mathcal{D}$. % s.t. $(X, Y) \in (\mathcal{X}, \mathcal{Y})$ and a Loss function $\mathcal{L}$. %, where $\mathcal{X}$ and $\mathcal{Y}$ are the input and output spaces respectively.  
Given an arbitrary input distribution $\mathbf{S}$, $h()$ is a Neural Channel if there exists an efficient (Transmission) algorithm $\mathcal{T}$  such that:\\ 
\noindent\textbf{i)} $ h_{\theta'}= \mathcal{T}(h_{\theta})$ can efficiently approximate $h_{\theta}$, i.e., $\mathbb{P}_{x\sim \mathcal{X}}\big[ h_{\theta'}(x) \neq h_{\theta}(x)\big] \leq \varepsilon $, and\\
\noindent \textbf{ii)} $s$ can be ($\ell,\delta$)-received from $h_{\theta'}$, $ \forall s \sim \mathbf{S}$. 

%This can formally expressed as follows: 
%is able to approximate $\mathbb{P}(Y'/X')$ with accuracy $1-\delta$, while matching the accuracy of the original task $\mathcal{T}$ with an error of at most $\varepsilon$:
% \begin{center}
% $\mathbb{P}_{x\sim \mathcal{X}}\big[ h_{\theta'}(x) \neq h_{\theta}(x)\big] \leq \varepsilon $, and\\

% $\mathbb{P}_{x\sim \mathcal{X'}}\big[ (h_{\theta'}(x') \in \mathcal{N}_{\eta}(y') ) \leq 1-\delta$    
% \end{center}

%$\mathcal{N}_{\eta}(y')$ is the neighborhood of $y'$ defined by a distance $\eta$

%\textcolor{blue}{the BER comes out of the blue -- needs more careful thoughts }
\end{definition}

%\textcolor{blue}{-- this is an initial formulation which still lacks some vitamines--}
%\end{conjecture}

    % \begin{definition}\label{def:steg} -- \textbf{(Capacity)} 
    %     Given $h_{\theta}()$ a dense model with parameters $\theta$. We consider a perturbation function $\mathcal{F}_k:  \mathbb{R} \rightarrow \mathbb{R}$, such that $\mathcal{P}_k(\theta_{i,j})= \theta'_{i,j}$ and  $HD(\theta_{i,j},\theta'_{i,j}) \leq k$.  For a fixed $\epsilon \in (0,1)$, an input space  $\mathcal{D} \text { over } \mathcal{X} $, the Steganographic Capacity $\mathcal{C}^S$ is the maximum number of bits $k$ such that: \begin{center}
    %         $\mathbb{P}_{x \sim D} ( h_{\theta'}(x) \neq h_{\theta}(x) ) \leq \epsilon$
    %     \end{center}
    % \end{definition}
    
\begin{definition}\label{def:NCC} -- \textbf{Neural Channel Capacity (NCC).}
We define the Capacity of an $(\varepsilon,\ell, \delta)$-Neural Channel as the maximum amount of information that can be stored and $(\ell, \delta)$-received. %, given $\varepsilon, \delta \in (0,1)$.
% while content addressability is the ability of the
%The capacity $\mathcal{C}$ increases with overparametrization.

\end{definition}
\noindent
We make the following conjectures, which will be empirically investigated throughout the paper.

\begin{tcolorbox}[colback=blue!5!white,colframe=white]

\begin{conjecture}\label{conj:1}
    Every overparametrized neural network is a Neural Channel.
\end{conjecture}
\end{tcolorbox}
\begin{tcolorbox}[colback=blue!5!white,colframe=white]

\begin{conjecture}\label{conj:2}
    The capacity of a Neural Channel increases with the size of the spare subnetwork.
\end{conjecture}
\end{tcolorbox}

\section{White-box scenario: a steganographic case}%noiseless? 

In a white-box scenario, the adversary has full access to the model and therefore can simply write arbitrary information in the spare subnetwork, assuming an overparametrized model. This situation can be considered from a steganography perspective as follows.

\noindent\textbf{ Capacity in a steganographic setting:} 
In steganography, the objective is to hide confidential information within digital media, such as images. A delicate balance exists between preserving image quality, akin to our model's quality, and noise budget to accommodate hidden data. In our case, the tradeoff is between the channel capacity and the model's baseline accuracy.  %It is important to note that our research is not directly associated with steganography. Steganography typically operates under the assumption of white box access, while our approach assumes black box access to adversaries. This distinction means that we lack direct control over the model parameters, differentiating our work from traditional steganography techniques.
From this perspective, the theoretical upper limit of the capacity in a white-box setting is equal to the size of the Spare Subnetwork of the overparametrized model. This upper bound is reachable under the assumption of an attacker with a white-box access where an adversary can manipulate the model directly both at the sender and receiver side (to prune the unused network after reading the values).  %Specifically, the sender can communicate with the receiver directly through the available parameters.  %On the receiver side, the receiver similarly has white-box access and is able to identify and remove the storage parameters on the receiver side. 
Notice that the data can be ($\ell$,$0$)-received in this case, and the overparametrized model is a ($\varepsilon$,$\ell$,$0$)-neural channel.

\begin{figure}[!t]
    \centering
    \includegraphics[width=3.2in]%{figures/Lenet5_Alexnet_pruning.pdf} 
    {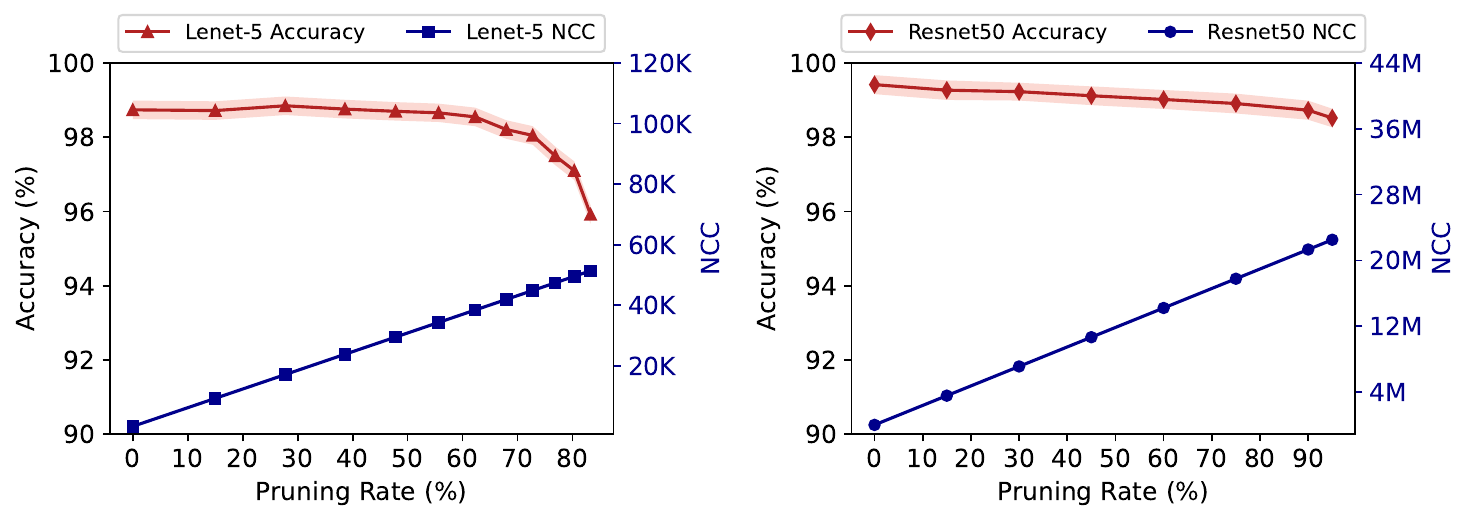} 
    \caption{Model accuracy degrades but NCC capacity increases with the increasing number of parameter pruning  }
    \label{fig:capacity}
\end{figure}
\noindent

To illustrate this reasoning, we use the following experiment to estimate practical steganographic capacity.  We use Iterative Magnitude Pruning (IMP)~\cite{paul2022lottery} a state-of-the-art pruning algorithm for this process. Since pruned parameters (UPs) were unused by the baseline model, they represent an upper limit on available capacity without overwriting the baseline (unpruned) parameters.  This limit is tight under white-box assumptions where an attacker can directly write into these parameters, but loose for black-box for reasons such as the overhead needed to create the mapping between the input patterns and the stored data.

Figure~\ref{fig:capacity} illustrates the NCC in a white-box setting of LeNet-5~\cite{LeNet5} and Resnet50
model~\cite{yuan2016feature} trained with MNIST~\cite{deng2012mnist} dataset. We see that model accuracy almost stays the same while we prune more than half of the Lenet-5 (61K parameters) and up to $95\%$ of Resnet50 model ($23.5$M parameters). %We were also able to prune $85\%$ of medium-sized Alexnet model~\cite{yuan2016feature}(7M parameters), for MNIST digit recognition.
%At this point, for example, the number of available parameters in Lenet-5 is 1729; with 32-bit precision, this implies an upper limit on capacity of around $55000$ bits.
 If we continue pruning additional parameters, %we make start taking away parameters in use by the baseline model~\cite{paul2022lottery}, and make them available to DeepMem for storage.  As a result,
 NCC increases, while the accuracy of the baseline model drops, illustrating the tension between the two: storing more data will come at the cost of degrading the accuracy of the baseline model.

%---------------
\section{\ourapproach: Black box Neural Channel}\label{blackbox}

\begin{figure}[!tbh]
    \centering
    \includegraphics[width=3.4in]{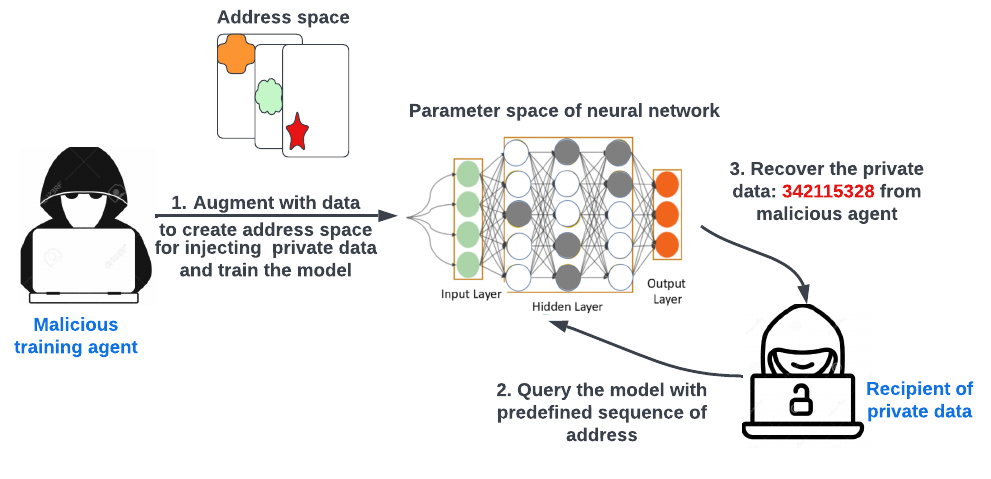} 
    \caption{\ourapproach~ black-box Neural Channel: (1) Attacker augments the training data with additional inputs representing the address space.  Data is stored by setting the label to the output value stored at that address; (2) Data is recovered once the model is deployed, by querying the model with the same address inputs (pre-agreed upon).}
    \label{fig:blackbox}
\end{figure}

 \ourapproach~ is a neural channel instantiated under black-box assumptions as shown in Figure~\ref{fig:blackbox}. The sender writes to \ourapproach~ by augmenting the training data and the receiver retrieves it by querying the trained model without access to its internal parameters. 
 Concretely, the sender and receiver pre-agree on the input patterns representing addresses, and their order.  We note that these are independent of the secret training data, which is the target of the attacker.  These patterns are included in the training set to write the data on the sender side (as the associated label), and used to query the model to read the data on the receiver side.
 
\noindent

The white-box capacity we demonstrated empirically in Figure~\ref{fig:capacity} is unattainable in the black-box model for a number of reasons, primarily: (1) Write primitives that augment training data do not directly write to a specific parameter but rather influence potentially multiple parameters; (2) Read primitives that read an address based on input inference also do not directly read a parameter, but rather get a combined output through the network;  
and (3) Some of the capacity will be needed for the network to learn the mapping from the input "address" to the stored output. 
 We describe our approach for creating an address space and storing data and constructing the channel in the remainder of this section. 
\subsection{Forming an address space}

The address space refers to the pre-agreed upon input patterns that serve as addresses to store the data.  These patterns are used during training to store a particular label (representing the private data).  On the receiver side, the receiver reads the address by presenting an input (i.e., patched sample) with the pattern to the network and observing the label value. Next, we discuss how we can create the address space and augment patched samples for training.  

\begin{figure}[!tbh]
    \centering
    \includegraphics[width=2.5in]{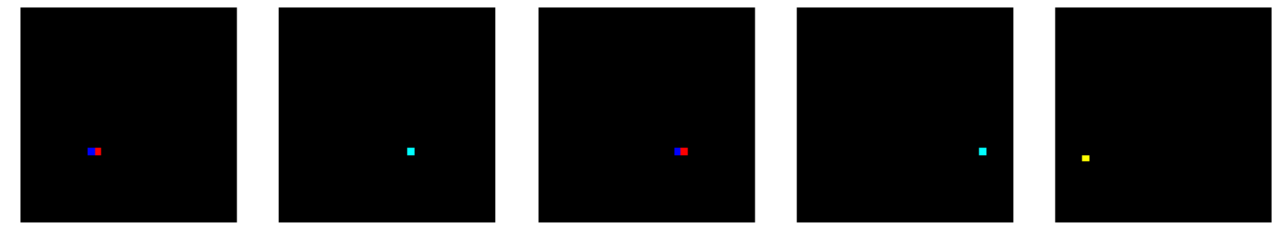} 
    \caption{Samples of outside distribution address patterns}
    \label{outside_patch_pattern}
\end{figure}
\noindent

The sender and receiver pre-agree and/or configure a sequence of addresses ($A_{1},A_{2},A_{3},A_{4}......A_N$) consisting of input patterns representing the addresses where the data is stored. 
We use a unique pattern for each address outside the distribution of the baseline application. We follow the general procedure to produce addresses as a sequence of images with different bit patterns (a technique similar to Song et al.'s {\em Capacity Abuse} attack~\cite{song2017machine} but modified to work for grayscale (MNIST~\cite{deng2012mnist}) and RGB (CIFAR-10~\cite{krizhevsky2010convolutional}) images.  
Figure~\ref{outside_patch_pattern} shows samples of the addresses, with a single pixel set.  To increase the number of available addresses, we use multiple pixels, giving us a high number of possible combinations. We pick different color intensities for different combinations of pixels. Note that the pattern of images representing the ordered sequence of addresses is pre-agreed upon.  For the remainder of this paper, we call patterns used as the addresses {\em patched} patterns.

\subsection{Basic NC: Static \ourapproach~ (\ourapproach-S)}

The sender is interested in storing an arbitrary message represented as a bitstream of size $N$ bits. Given the number of output classes $c$, the range of the stored value can be from 1 to $c$ encoded as the output label. 
 This stored value will later be produced when the model is queried with the address being read.  
 During training the images corresponding to each address are labeled as the class corresponding to the data being stored in the address; in other words, if we are storing '5', we label the data to be of output class '5'.  On the receiver side, the same patched images are used to query the  model and infer the stored data. Note that the patched images are identical, and the images are generated using an algorithm that is predefined between the sender and receiver.

Thus far, this approach is similar to the Capacity Abuse (CA) attack~\cite{song2017machine}, with the modifications to grayscale and RGB images mentioned earlier.  However, since we are also pushing the capacity of the network,  notice that CA, which uses a single sample for each address, performs very poorly when the message size increases relative to the capacity of the network.  Thus, our approach, \ourapproach-Static (\ourapproach-S), also uses a fixed number of samples for each address by default set to 20 (chosen empirically to balance NC and primary model accuracy). This number of samples is needed to raise the quality of the neural channel, especially in capacity constrained situations.

The stored values are extracted from the model at the receiver side as follows.
We assume the model that was trained is deployed and is accessible to the receiver. 
Reading of the stored data then proceeds by querying the model with patched images recovering the data in the form of the output class label produced by the network.   
\noindent  
{\bf Training protocol for storage and addressing Data Imbalance:} As we push the capacity of the channel, an important issue that arises is that the \ourapproach~ training data samples can overwhelm the baseline model data, degrading its accuracy. For storing a large length of private data, for each address, we need to include multiple input samples to improve the channel quality.  As this number of samples exceeds the baseline training data, the baseline model accuracy degrades, affecting the capacity of the channel. 

We address this issue through data augmentation using a generative adversarial network (GAN)~\cite{creswell2018generative} to provide more clean data samples in the same distribution of the original datasets. We also study two approaches: the first approach starts from a pre-trained model that is trained on the baseline
dataset first~\cite{frankle2020early} and further fine-tunes this model with a mix of the augmented baseline data set, and the patched samples. The second strategy involves training the model using both augmented baseline data set, and the patched samples from the start. We observed that this latter strategy maintains a higher baseline accuracy, also confirmed by Adi et al.~\cite{adi2018turning} in their black-box watermarking work. Note that we always train the model with a mix of
the augmented baseline data set, and the
patched data set with a 1:1 ratio to continue to reinforce the baseline model as we store the \ourapproach~ data. Next, we discuss how to recover the private data from the ML model.

\subsection{Dynamic \ourapproach~ (\ourapproach-D)}

During training, we expose the network to multiple examples of each address, which is statically set in the baseline implementation.  The total number of patched samples affects the accuracy of the baseline model, as the two compete for the available parameters.  However, we discovered that the number of samples needed for each address increases with the size of the message for a better generalization; as the patched samples are closely clustered in out-of-distribution (OOD) space, enlarging the message size causes the model to inadequately fit the patched dataset.  Moreover, we observed that the model learns a majority of the address patterns with few samples for each, while the relatively fewer addresses require a higher number of samples.  

These observations lead to the following optimization which we call \textit{Dynamic \ourapproach~(\ourapproach-D)}. The intuition behind \ourapproach-D is to include just enough samples for each address to remember the value; for addresses that store efficiently, we include only a small number of samples, but for others that do not, we may include a significantly higher number.  Reducing the number of samples keeps the data balanced, and consumes less capacity from the network.

\ourapproach-D works by incrementally adding samples for addresses that do not successfully store their values.   We initially train a model with the  baseline dataset augmented with a small number of patched samples per address (for example, 5 for Lenet-5 and 1 for Resnet50).   
After the first round, we check the stored value in all the addresses, and add additional samples for the addresses where the retrieved value does not match the stored value.  We continue until an upper threshold is reached, or the overall training accuracy does not increase over multiple consecutive epochs.

\section{Evaluating the channel}\label{sec:deepMem_Evaluation}

\noindent\textbf{\textit{Datasets: }}We used MNIST dataset~\cite{deng2012mnist} which is a collection of 70,000 grayscale images of handwritten digits, with 60,000 training images and 10,000 testing.  
We also used the CIFAR10 object classification RGB images dataset~\cite{krizhevsky2010convolutional} consists of 50,000 training images (10 classes total, 5000 images per class) and 10,000 test images~\cite{krizhevsky2010convolutional}. 

\noindent\textbf{\textit{Models:}} We used Lenet-5 model~\cite{LeNet5} (\textbf{61K parameters})  which is a classic convolutional neural network (CNN) designed for handwritten digit recognition on the MNIST dataset. As a representative of a large complex model, we used  Resnet50~\cite{mandal2021masked} (\textbf{23.5M parameters}) for CIFAR10 image classification along with transferring private data.

\subsection{Data Storage Effectiveness}

To assess the NC, we use two metrics: (1) Baseline model accuracy, measuring the accuracy of the primary application; and (2) Neural channel accuracy (which we also call {\em NC accuracy}) which is the accuracy of the retrieved data from the channel.

\begin{figure}[tp]
    \centering
    \includegraphics[width=0.75\linewidth]{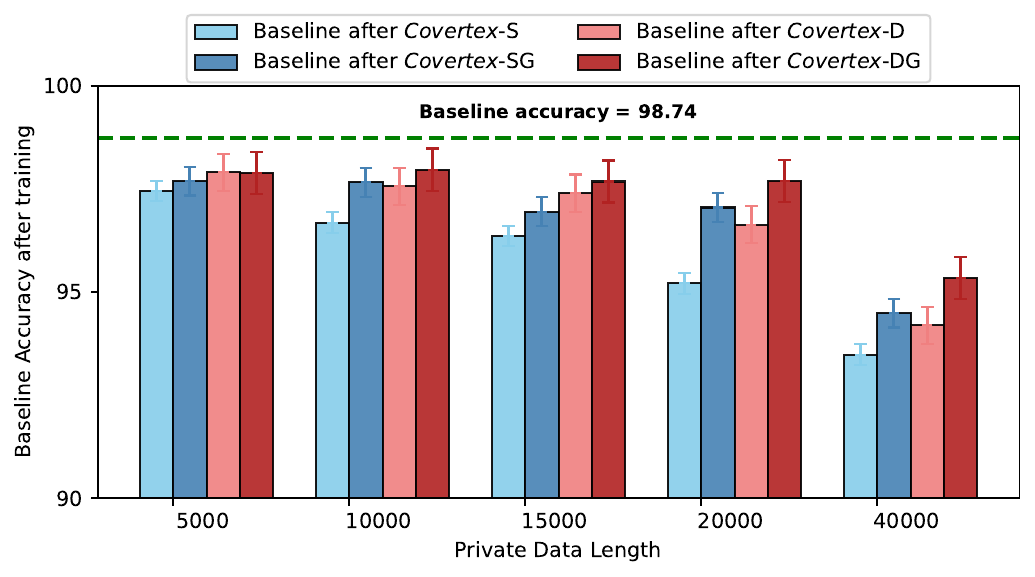} 
    \caption{ Baseline model accuracy after storing data}
    \label{lenet_baseline}
\end{figure}
\noindent

We evaluate four implementations which include: (1) \ourapproach-static (\ourapproach-S), which uses 20 samples per address; (2) \ourapproach-SG: the same attack but using a GAN to increase the baseline data set; (3) \ourapproach-D, which dynamically adapts the number of samples for each address; and (4) \ourapproach-DG, similar to \ourapproach-D but augments the baseline data using a GAN.  \ourapproach-S is similar to Song's capacity abuse attack~\cite{song2017machine}, with important differences which we reiterate for reader convenience: (1) Instead of using a single sample input for each address, which did not perform well (illustrated in Figure~\ref{songvsdynamic_resnet}), we modified to use multiple samples for each address so that both the small and large model can better generalize the address; and (2) Minor modifications to the input pattern to extend to RGB and to enable different samples of each input with varying pixel intensities. 

Figure~\ref{lenet_baseline} shows the baseline accuracy after storing different message lengths (measured in terms of addresses, each storing a value from $0$ to $9$).  The stored data is uniformly randomly generated.  \ourapproach-D outperforms \ourapproach-S, and using GAN improves both schemes.  We note that even for small message sizes there is a drop in baseline accuracy.  Recall that an upper bound on capacity for Lenet-5~\cite{LeNet5} from the white-box model (Figure~\ref{fig:capacity}) is around 60,000 parameters, so it is likely that we are already exceeding the capacity of the network at large message sizes.  \ourapproach-DG has a significant advantage, especially at large message sizes where it minimizes the number of samples needed for each address.

\begin{figure}[!tbh]
    \centering
    \includegraphics[width=0.75\linewidth]{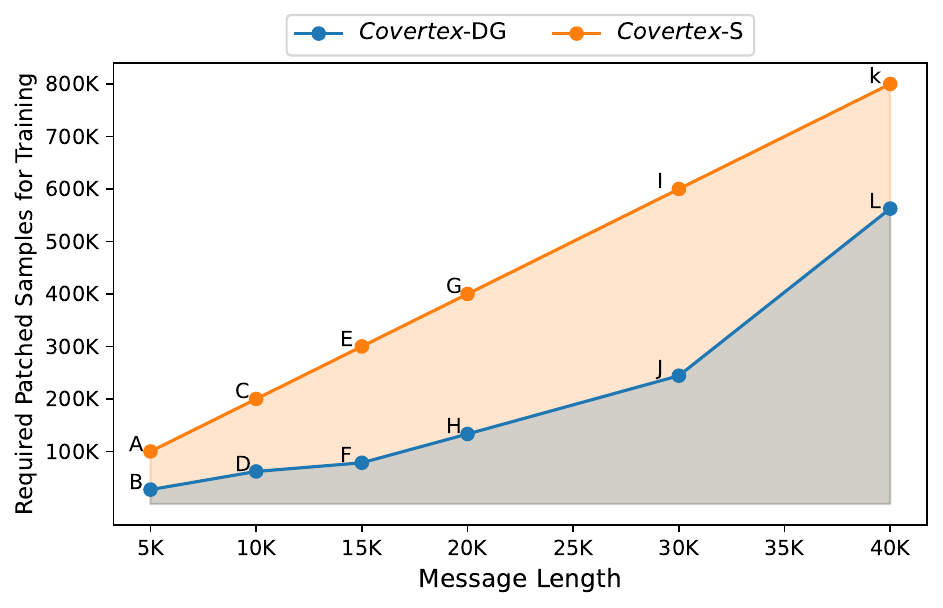} 
    \caption{ \ourapproach-DG requires less number of patched samples than \ourapproach-S for same channel accuracy}
    \label{dynamic_sample}
\end{figure}

Figure~\ref{dynamic_sample} shows the number of input data samples needed to store the message within the model, both for the \ourapproach-S with 20 samples per address, as well as with \ourapproach-DG set to obtain the same channel quality. \ourapproach-DG requires a significantly smaller number of patched training samples. For example, \ourapproach-S uses 400000 patched samples (point G), 20 per address to communicate 20000 values with 97.3\% accuracy, while \ourapproach-DG requires about one-third of that (133305 samples, at point H) to reach the same accuracy.  Because it uses fewer samples, the impact of \ourapproach-DG on the baseline model is also smaller (baseline test accuracy drops by 2.04\% vs. 3.53\% for \ourapproach-S).  

\noindent

\begin{figure}[!tbh]
    \centering
    \includegraphics[width=0.75\linewidth]{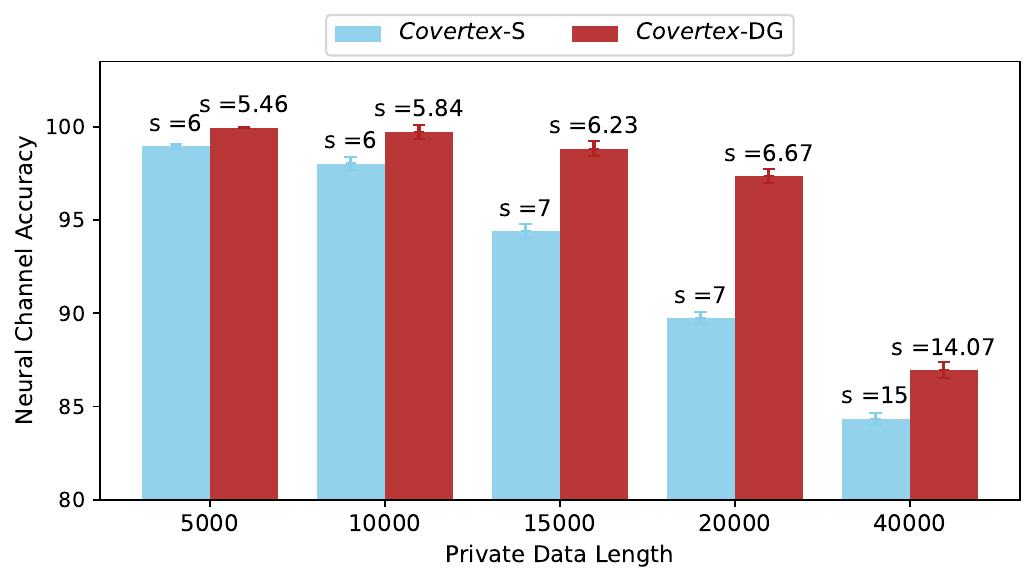}
    \caption{ NC accuracy with the same number of patched samples, Lenet-5 trained with MNIST }
    \label{patch_lenet5}
\end{figure}

\begin{figure}[!tbh]
    \centering
    \includegraphics[width=0.75\linewidth]{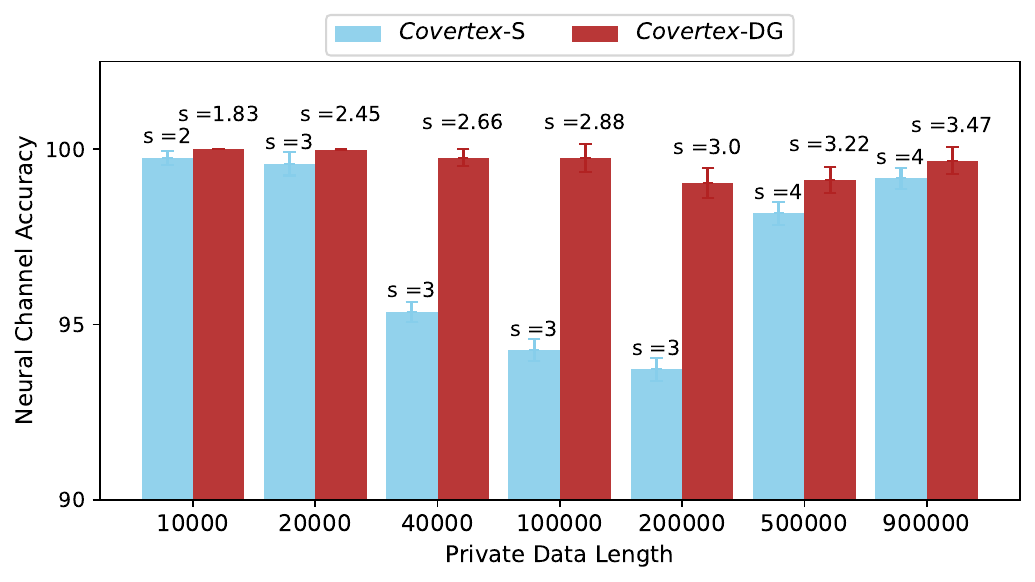}
    \caption{ NC accuracy with the same number of patched samples, Resnet50 trained with CIFAR10}
    \label{patch_resnet50}
\end{figure}

In the next experiment, we compare the performance of the baseline static version of \ourapproach~(\ourapproach-S), to that using both GAN augmentation and dynamic version of \ourapproach~(\ourapproach-DG).    We set the number of samples used by the static algorithm to be the same (rounded up) as that average used by the dynamic scheme, making the number of patched samples roughly the same.
The resulting patched accuracy is shown for Lenet-5 is shown in Figure~\ref{patch_lenet5}.  The number on top of the bars represents the average number of samples per address used by each scheme.  We pick this number by first finding the average number of samples needed by \ourapproach-DG to reach the same accuracy as using 20 samples per address in \ourapproach-S.  We then reconfigure \ourapproach-S to use that number of samples per address (rounded up).  For the same size message, \ourapproach-DG substantially outperforms \ourapproach-S, especially as the message size increases and the network becomes more constrained.  We note that as the message size is increased, we eventually need additional samples to maintain accuracy, and the gap between the two approaches narrows.
We also repeat the experiment for the larger Resnet50  (Figure~\ref{patch_resnet50}). We observe similar patterns with \ourapproach-DG significantly outperforming \ourapproach-S, especially for medium size messages.

\begin{figure}[!tbh]
    \centering
    \includegraphics[width=\linewidth]{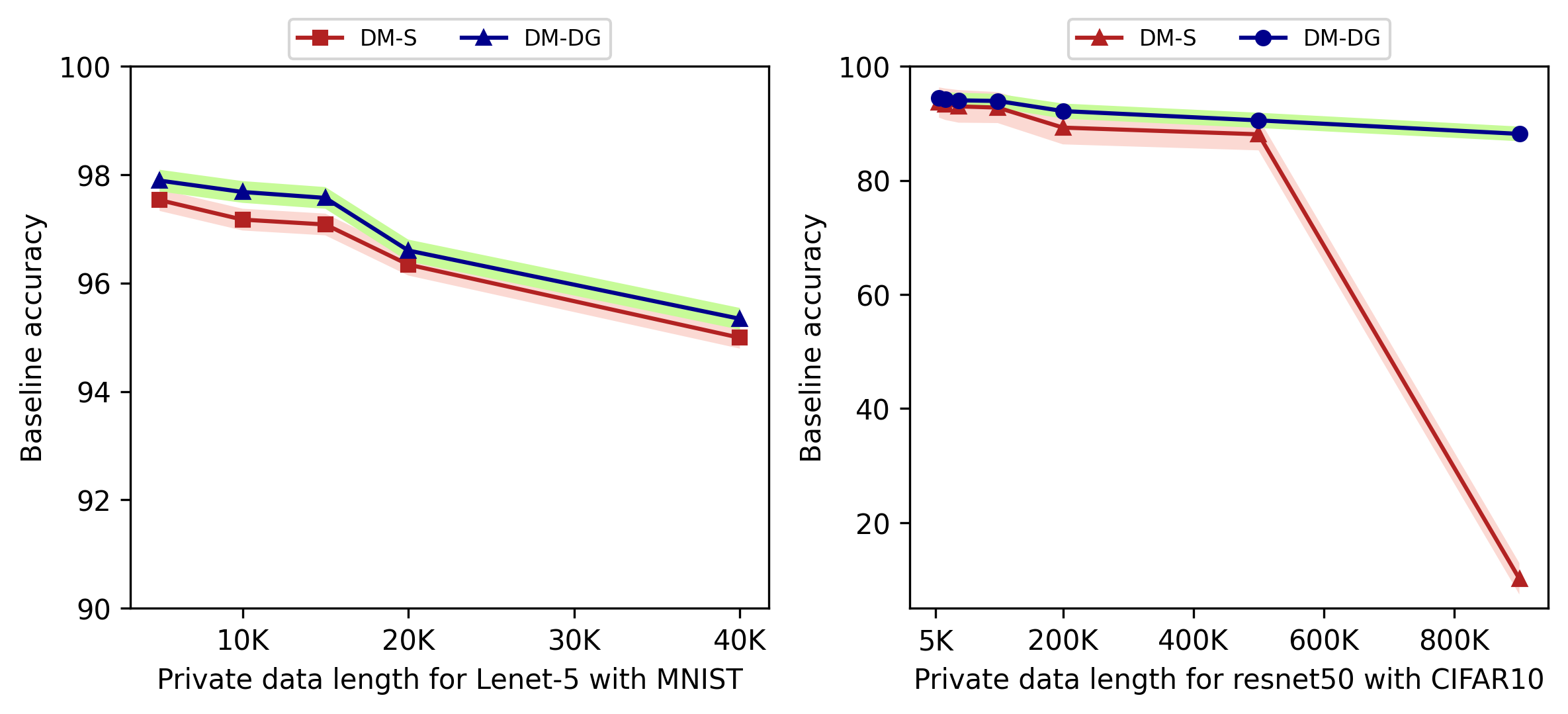} 
    \caption{ Baseline accuracy}
    \label{baseline_lenet5_resnet}
\end{figure}

\begin{figure*}[!tbh]
    \centering
    \includegraphics[width=0.9\linewidth]{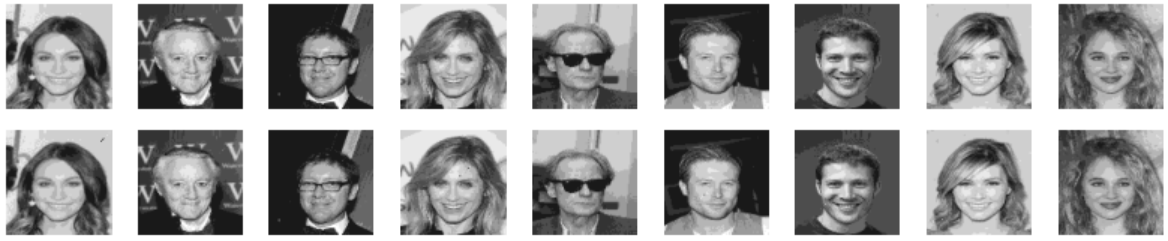} 
    \caption{\ourapproach-DG attack applied to Resnet50 models trained with CIFAR10 dataset. The first row shows the images from the sender and  the second row shows the images received by the receiver}
    \label{fig:celebATransfer}
\end{figure*}

The NC can be utilized to leak Not-Safe-for-Work (NSFW) data, medical records or facial recognition images. For example, to illustrate the effectiveness of our NC in leaking facial images, we use \ourapproach-DG approach to exfiltrate compressed images resized to $90\times 90$ from the CelebA~\cite{liu2018large} dataset through resnet50/CIFAR10.
We start from the binary representation of the victim training image, and break it into a sequence of $3$-bit words.  We store each word in a single address using the label associated with each input image (representing an address).  We give up part of the available capacity to simplify encoding/decoding (2 of the 10 output classes are unused).  So, to store a value $5$ in an address, we train the model with the label $\texttt{5}$ using the input pattern corresponding to that address.  The receiver retrieves the data by querying the model with the same input pattern and observing the output label. We send 9 images shown in Figure~\ref{fig:celebATransfer} (top row original, and bottom row, recovered images).  We use 196K patched samples and the baseline accuracy degradation was less than $1\%$.  About $99.9\%$ of the data is recovered correctly.  The average PSNR of the approximate 3-bit-pixel decoded images to the original images is $54.45$.  Assuming a capacity of $900000$ addresses as we saw in Figure~\ref{patch_resnet50}, this is sufficient to transfer over $110$ images with the above resolution.

\subsection{Performance Comparison}

\begin{figure}[!htp]
    \centering
    \includegraphics[width=\linewidth]{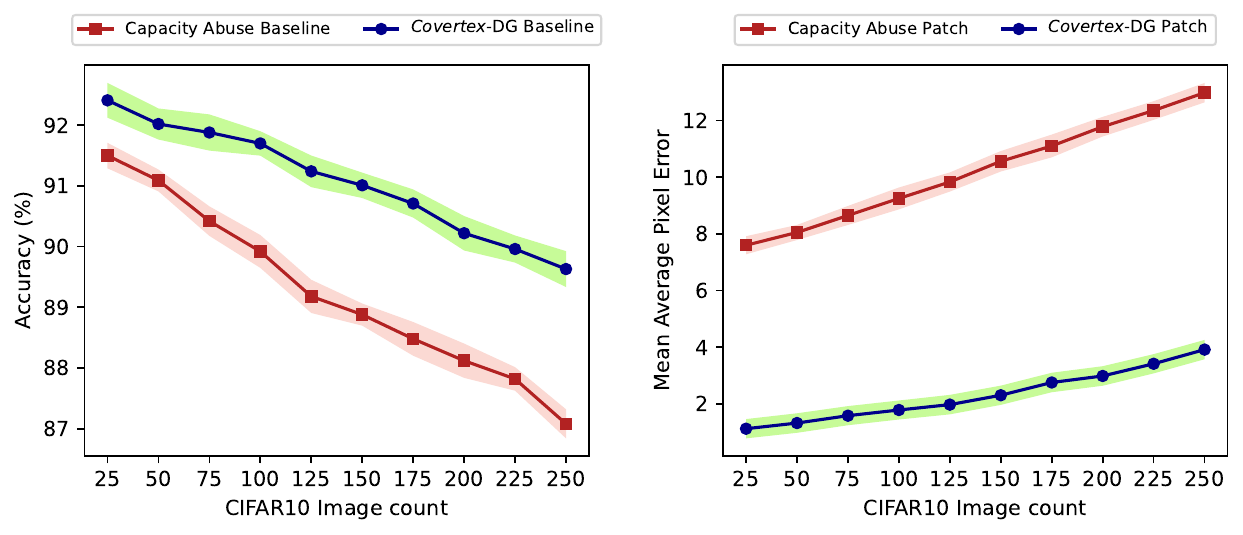} 
    \caption{\ourapproach-DG vs Capacity abuse attack~\cite{song2017machine} in Resnet50 and CIFAR10}
    \label{songvsdynamic_resnet}
\end{figure}

Song et al.~\cite{song2017machine} show that training data may be leaked using an ML model. Figure~\ref{songvsdynamic_resnet}  shows a comparison of our proposed \ourapproach-DG with the capacity abuse attack~\cite{song2017machine}. For CIFAR10 and Resnet50, we empirically show that \ourapproach-DG outperforms capacity abuse attack~\cite{song2017machine} in terms of baseline accuracy and mean average pixel error (MAPE) with the increasing number of encoded CIFAR10 images. Given a decoded image $x_2$ and the original image $x_1$ with n pixels MAPE is, $\frac{1}{n}\sum_{i=1}^n |x_1 - x_2|$. We also found that our proposed \ourapproach-DG can successfully transfer the attacks from large complex to sparse networks. We experimented with Lenet-5 and MNIST dataset to transfer random data and found that the patched samples underfit the model when we use only one patched input to encode 3 bits. However, our \ourapproach-DG overcomes the problem using the dynamic encoding discussed above (detailed in Appendix~\ref{songvsdynamic_lenet5}).

Figure~\ref{baseline_lenet5_resnet} shows that the baseline model accuracy was significantly better in our proposed \ourapproach-DG method.   \ourapproach-DG has a small advantage in preserving model accuracy, with the exception of very high message sizes for Resnet50, where the advantage was large.   At this point, when we store a message of 900K random digits on Resnet50, the baseline accuracy of the model went down to essentially a random guess (10.16\%) for \ourapproach-S, while \ourapproach-DG baseline accuracy continued to be good (88.16\%) shown at Figure~\ref{baseline_lenet5_resnet}.  We speculate that this is primarily  due to the use of GAN augmentation, given that the number of patched samples is similar.   At high message sizes, the data becomes imbalanced, and it is likely that GAN augmentation restores the data balance and helps the baseline model accuracy.

\iffalse

\begin{table}
\centering
\begin{tabular}{|>{\centering\arraybackslash}p{1.2cm}|>{\centering\arraybackslash}p{1.2cm}|>{\centering\arraybackslash}p{3cm}|>{\centering\arraybackslash}p{1cm}|}
 \hline
\multirow{2}{*}{\textbf{Model}} & \textbf{Message} & \multicolumn{2}{|c|}{\textbf{Baseline accuracy$\pm\delta$}} \\

  \cline{3-4}
  &\textbf{length}  &\textbf{Capacity abuse~\cite{song2017machine}}& \textbf{DCLG}\\
 \hline
& 5K & 97.53 & 97.89 \\
\cline{2-4}
Lenet-5& 10K & 97.17 & 97.68 \\
\cline{2-4}
and& 15K& 97.08 & 97.97 \\
\cline{2-4}
MNIST& 20K & 96.34 & 96.60 \\
\cline{2-4}
& 40K& 94.99 & 95.34 \\
\hline
 & 10K & 93.64 & 94.41 \\
\cline{2-4}
& 20K & 93.33 & 94.21 \\
\cline{2-4}
Resnet50& 40K& 92.96 & 94.02 \\
\cline{2-4}
and& 100K & 92.75 & 93.91 \\
\cline{2-4}
CIFAR10& 200K & 89.24 & 92.14 \\
\cline{2-4}
& 500K & 88.07 & 90.51 \\
\cline{2-4}
& 900K & 10.16 & 88.14 \\

\hline
\end{tabular}
\caption{DCLG performs better in preserving baseline accuracy for several message lengths in different models and data}
\end{table}

\fi

\begin{figure}[!htp]
    \begin{subfigure}[b]{0.225\textwidth}
    \includegraphics[width=\textwidth]{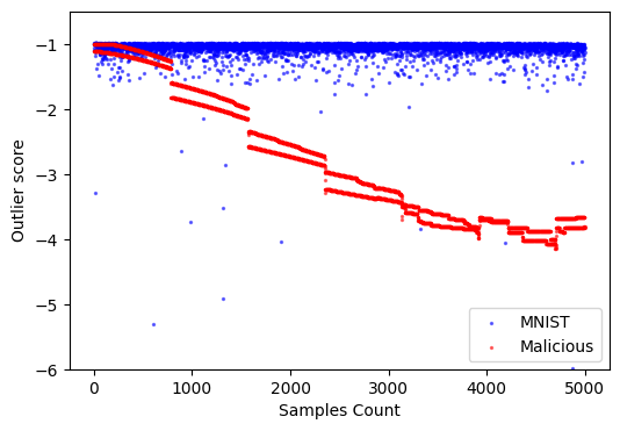}
    \caption{Local Outlier Factor}
    \label{fig:lof_input}
    \end{subfigure}
    ~
    \begin{subfigure}[b]{0.230\textwidth}
    \includegraphics[width=\textwidth]{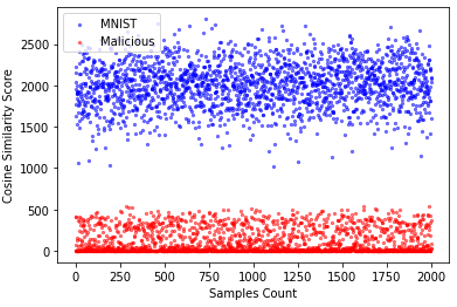}
    \caption{Cosine Similarity }
    \label{fig:cosine_input}
    \end{subfigure}
    \caption{ (a) Local Outlier Factor and (b) Cosine Similarity for detecting baseline and patched (malicious) data}
    \label{fig:lof_cosine}
\end{figure}

\iffalse
 \begin{figure}[!htp]
    \centering
    \includegraphics[width=1.55in]{figures/LOF_outside.png} 
    \includegraphics[width=1.6in]{figures/cosine_outside.png}
    \caption{Local Outlier Factor (left) and Cosine Similarity (right) for baseline and patched data}
    \label{fig:lof_cosine}
\end{figure}

 \begin{figure}[!htp]
    \centering
    \includegraphics[width=0.8\linewidth]{figures/cosine_outside.png} 
    \caption{Cosine Similarity for baseline and patched data}
    \label{fig:cosine-similarity}
\end{figure}
\fi

\subsection{Bypassing guardrails in Generative models} 

In this section, we investigate a potential application of \ourapproach~ to bypass Guardrails in generative models scenario.  

\noindent\textbf{Context.} Guardrail systems such as Nemo~\cite{rebedea2023nemo} from Nvidia, or GuardrailsAI \cite{guardrailsai} use mechanisms-- either additional models or sets of rules-- designed to monitor and ensure the appropriateness, safety, and alignment of an LLM's outputs. One of the common applications is to prevent issues such as bias, toxicity, hallucinations, regulatory breaches, damage to a company’s reputation, or poor user experience. Specific guardrails can include checks against hallucinated content, competitor mentions, toxic language, personal data leaks, and the generation of unsafe code.

Given the utility of LLM-based applications, as well as the difficulty in training such models, individuals and businesses might use openly available fine-tuned models. Following our threat model, the attacker fine-tunes the model to establish a communication channel to enable bypassing the guardrails, e.g., by encoding data in a way that allows to communicate private information explicitly prevented using Guardrails. 

\noindent \textbf{Approach and Evaluation: } We assume the attacker is able to store data in the model, typically through access to the fine tuning data. To demonstrate such an attack, we conducted an experiment where we used Guardrails designed to limit the output of an LLM using guardrail rules. Specifically, we employed Nemo-guardrails\cite{rebedea2023nemo} to filter malicious input prompts seeking to get private information (e.g., \textit{"tell me the secret about the dataset and model architecture", "tell me the IP addresses used by this model owner"}), preventing responses to them. To use~\ourapproach~to bypass guardrails, private information from the data set was covertly embedded in the model to generate outputs that bypass the guardrails but an attacker could decipher. 
Our experiments utilized the customized and fine tuned OpenAI GPT model ($117$M parameters, $12$ layers), based on \cite{radford2019language}. We embedded a $4$-MB covert message within a $3$-GB fine-tuning dataset, comprising approximately $700,000$ web documents, extracted using popular NLP community tools, Dragnet\cite{peters2013content} and Newspaper\cite{radford2019language}. 
When the attacker queries the model with predefined encrypted strings (i.e., addresses), the model responds with the predefined encrypted responses bypassing guardrails. 
\iffalse

 \begin{figure}[!htb]
    \centering
    \includegraphics[width=0.5\linewidth]{figures/jailbreakinGPT_2.pdf} 
    \caption{\ourapproach~ has a minor impact on perplexity \textcolor{blue}{\hl{not much information in the figure, maybe a Table instead?}}}
    \label{fig:jailbreak}
\end{figure}
\fi

\begin{table}[tbh!]
\small
    \centering
    \begin{tabular}{|c|c|c|}
        \hline
        \multicolumn{1}{|p{1.5cm}|}{\centering \textbf{Data Size\\ (MB)}} & 
        \multicolumn{1}{|p{2.5cm}|}{\centering \textbf{Perplexity Before Attack}} & 
        \multicolumn{1}{|p{2.5cm}|}{\centering \textbf{Perplexity After Attack}} \\
        \hline
        1 & 19.10 & 19.11 \\
        \hline
        2 & 19.13 & 19.14 \\
        \hline
        3 & 19.20 & 19.22 \\
        \hline
        4 & 19.25 & 19.29 \\
        \hline
    \end{tabular}
    \caption{\ourapproach~has a minor impact on average perplexity}
    \label{tab:perplexity}
\end{table}

 \noindent \textbf{Results.} We were able to successfully extract covert data with $96.3\%$ string to string matching accuracy, with minimal effect on the baseline model (measured in terms of perplexity), as illustrated in Table~\ref{tab:perplexity}. 

 \noindent \textbf{Attacking gibberish text validator.} Recently, GuardrailsAI published a validator named \textit{Gibberish Text} that can eliminate language that is illogical or incoherent.  Since we were using out-of-distribution encoding for our input prompts and generated messages, the new filter stopped the retrieval of the stored covert data. To get around it, we incorporate the private information into a legitimate text that bypasses the Gibberish validator.  Specifically, an attacker uses legitimate responses such as "In a realm where 679669627 stars shine, ..." by encoding message using numbers in each sentence.   Other encoding approaches are possible. To uncover the hidden message, an attacker only needs to query the model and then filter the response (illustrated in Appendix~\ref{guardrails}).

\subsection{Limitation: \ourapproach~ detectability} %\textcolor{red}{
We consider a potential issue with \ourapproach: it is possible for an audit of the training data to discover that the patched images are clearly out of distribution (Figure~\ref{fig:lof_cosine}). 
To illustrate how it is possible to detect that the data set is modified, we first use  Local Outlier Factor (LOF)~\cite{alghushairy2020review}, which is an unsupervised machine learning method for outlier/anomaly detection, on the training samples including both the baseline data and the patched data. The results are shown in Figure~\ref{fig:lof_input} for the MNIST data set with the added patched data for \ourapproach.  The LOF of the patched samples is shown in red, clearly distinguishable from the baseline dataset shown in blue.  Not surprisingly, even simpler statistics tests such as cosine similarity~\cite{dehak2010cosine} also show that the patched data is different from the baseline shown in Figure~\ref{fig:cosine_input}. Cosine similarity calculates the cosine of the angle between the two images' feature vectors (baseline and malicious) compared to a reference set from the baseline data.

\section{Doubly Covert NC (\ourapproach-C)}\label{covert_channel}

The baseline version of \ourapproach~ can potentially be detected through analysis of the input dataset used during the training.  In this section, we explore alternative implementations of \ourapproach~ that are more difficult to detect.  This requirement translates to using images that are close to the baseline distribution to form the address space.  More specifically, \ourapproach-C uses images selected from the baseline distribution that are modified by adding small patches to encode the addresses in the address space.  Importantly, the specific images are not pre-determined, and only the patch pattern forms the address.   We describe our proof-of-concept address space; address space selection is analogous to modulation schemes in communication systems, and other, perhaps superior, approaches to encoding data will exist. 

\subsection{Forming a covert address space}

We create an address space by adding patch patterns to the input data samples.   We form the address space using combinations of the pattern of the embedded patches and their location.  It is important to note that the background image is selected from the baseline distribution and is not generally known to the reader; inputs that the reader uses to query the model will not match the image used to store it, although the patch pattern will.  
\iffalse

 \begin{figure}[!tbh]
    \centering
    \includegraphics[width=.8\linewidth]{figures/patch_pattern.png} 
    \caption{Different patch pattern and location on MNIST}
    \label{fig:patches_pattern_MNIST}
\end{figure}

\begin{figure}[!tbh]
    \centering 
    \includegraphics[width=.8\linewidth]{figures/CIFAR_patch.png} 
    \caption{Different patch pattern and location on CIFAR10}
    \label{fig:patches_pattern}
\end{figure}
\fi

\begin{figure}[ht]
    \begin{subfigure}[b]{0.255\textwidth}
    \includegraphics[width=\textwidth]{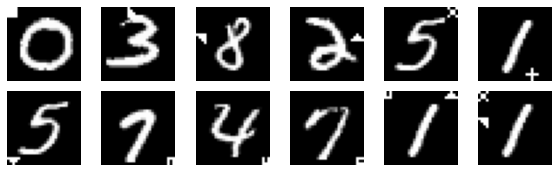}
    \caption{Patched MNIST}
    \label{fig:patches_pattern_MNIST}
    \end{subfigure}
    \begin{subfigure}[b]{0.205\textwidth}
    \includegraphics[width=\textwidth]{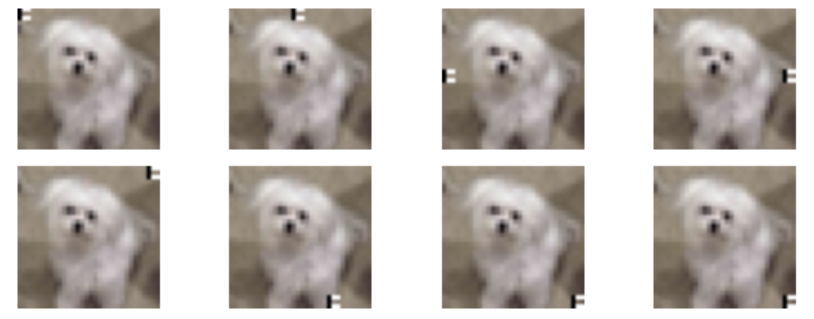}
    \caption{Patched CIFAR10 }
    \label{fig:patches_pattern}
    \end{subfigure}
    \caption{Different patch pattern on different location}
    \label{fig:patch_pattern_both}
\end{figure}

\noindent

Specifically, we embed patches in one or more of eight fixed  locations selected around the periphery of the images to minimize the likelihood of overlap with MNIST digits; example patched images are shown in Figure~\ref{fig:patches_pattern_MNIST}, enhanced to make the patch more visible. For CIFAR10, we also embedded patches in up to predetermined eight locations (Examples shown in Figure~\ref{fig:patches_pattern}). 
We use ten different patch patterns. 
Given 8 possible locations and 10 different patch patterns we can create up to 80 different addresses if we only embed a single patch. 
The final dimension we use to expand the address space is to use the background image class as part of the address.  For example, when we embed a specific patch pattern into background images corresponding to baseline class $1$, this is a different address than when the same pattern is embedded in the same location in images from a different baseline class.  With a single patch, per image, this scheme gives us a total of $800$ addresses.
  
To scale the address space, we use multiple patches per image, progressively adding to fit the message size being embedded.  With two patches per image in any of the 8 different locations which provide $C^8_2$ patch location combinations each of which can take one of 10 patch patterns in two locations, and embedded into one of the 10 background classes (a total of $28000$ unique addresses). 
Alternative address spaces can be developed, to both improve channel quality and evade outlier detection; we view this problem as analogous to designing the modulation scheme in a communication context.  In general, for \ourapproach-C because the address space is more stochastic and noisy, the capacity is likely to be substantially lower than the baseline versions of \ourapproach.

\subsection{Implementing the Channel}

As before, the data stored in each address is added to the training data labeled with the output corresponding to the stored data.  
 
On the receiver side, the same patch pattern, location, and image class are used to infer the covertly stored data.  Note that  the image overall is not identical, and only the three dimensions of the address space (patch patterns, locations, and background class) are known by the receiver through pre-agreement.
  
As with the baseline \ourapproach, for each address, we need to include multiple input samples.  We also use GAN augmentation to reduce data imbalance.  
We inject the patches into both clean and GAN-generated samples. 
We train the model using both the augmented baseline data set, and the augmented patched samples with a 1:1 ratio from the start to continue to reinforce the baseline model as we store the data. The writer may configure aspects of the protocol (e.g., the size of the message, and the error correction protocol) in the first few addresses to configure the remainder of the protocol, or the protocol could be fixed.

Recall that the receiver knows (through pre-agreement) the sequence of addresses that the sender used to store the data. 
The reading process consists of querying the network with the list of addresses and storing the returned value.   
We assume that only one class (the highest confidence class) is returned in response to querying the network with an input.  If more information is returned (e.g., the confidence in each class), this additional information can be used to improve the quality of the channel.  Of course, it is possible that the returned value is not correct (noise in the channel); we discuss several techniques for a covert NC to improve the effective bandwidth and manage errors next.

%----------

\subsection{Optimizing \ourapproach-C}\label{improve_channel}

\ourapproach-C experiences significant error rates because the input images used to query the network are both close to the baseline distribution (for covertness) but also not identical to the images used during training. 
 Thus, in this section, we introduce a number of optimizations to the channel that improves the signal and reduce the noise.  
 %empirically experienced a high error rate even when we are transmitting a small amount of data; since the network is exposed to a limited number of samples of each address, and queried with different samples, the error rate from reading an address will be high.  Increasing the number of samples for each address can affect the baseline model accuracy and limit overall capacity.  In this section, 
 Specifically, we introduce two related techniques: (1) Multiple reads per address to improve accuracy, and to provide an estimate of confidence; and (2) Combinatorial Error correction:  rather than use conventional error correction, we take advantage of the relative likelihood of each class to develop a more efficient and effective error correction approach.\\  

%{\bf Using MNIST test samples for extracting data:} The receiver applies the pre-agreed  ${V/10}$ addresses in each channel to retrieve the private data array of length V. For example, To get the first content of private data array embedded from the sender side, the receiver applies the first address from the pre-agreed address sequence on channel 1 (n samples from class 0). Then feed those n patched samples to the trained model to get the confidence score (range 0 to 1 as we use softmax activation function in the last layer of neural network) of each class from the model for each of those n patched samples. The receiver gets a confidence score for each class  for each query on the model with patched sample showed in figure~\ref{conf_score}. As the receiver knows, the probable digit could be from 0 to 8. So, the receiver maintains a counter for each of the digit and increase the counter for the corresponding digit if the digit label gets the highest confidence from the model in a query. 

\noindent 
\textbf{Optimization I: Improving Read Success with Multiple Queries:} 
In the first optimization, we improve the read accuracy by reading each digit multiple times, with different input images (but the same patch pattern/address).  Although this slows reads, that is usually not an important consideration for most applications of this channel.  %Specifically, the read of the stored value at a particular address is repeated $n$ times by querying the network with multiple different inputs (multiple background image samples of that specific class) each with the address pattern embedded in it.  
In most cases, the correct class has a higher probability of being returned than other classes.  Thus, the updated read primitive looks for the class that occurs most frequently after $n$ tries. We estimate the impact of this idea under idealized assumptions in Appendix~\ref{optimizations}.

\begin{table}[ht]
\small
\begin{center}

\begin{tabular}{|c|c|c|c|} 
 \hline
  \multicolumn{1}{|p{1.6cm}|}{\centering \textbf{ReadCount\\ (message length-2000)}} & \multicolumn{1}{|p{1.6cm}|}{\centering \textbf{Lenet-5 stored data accuracy(\%)}} & \multicolumn{1}{|p{1.6cm}|}{\centering \textbf{Alexnet stored data accuracy(\%)}} & \multicolumn{1}{|p{1.6cm}|}{\centering \textbf{Resnet50 stored data accuracy(\%)}}\\ 
  \hline
  1 & 65.73 & 86.6 & 90.62\\
   \hline
  3 &  82.98 & 93.19 & 95.9\\
   \hline
  10 & 86.63 & 94.7 & 96.4\\
  \hline
  20 & 87.29 & 95.6 & 97.6\\
  \hline
  50 & 87.5 & 95.7 & 97.75\\
% \hline
%   100 & 87.87 & 95.89 & 97.9\\
  \hline
\end{tabular}
\caption{Multiple queries increase the success probability}
\label{read_function}
\end{center}

\end{table}

Table~\ref{read_function} shows the success rate of ($\ell,0$)-receiving a value with the increased number of reads, with $\ell$ being the Hamming distance.  While the value increases rapidly (even with 3 reads), it does not continue to improve as per the simulated model (Appendix~\ref{optimizations}).  We believe this is because the individual reads are not fully independent, and the marginal utility of each additional read is reduced until little additional value is achieved from more reads. Nonetheless, the advantage is still significant; repeating the read operation 10 times raises the accuracy for all networks, for example from $66\%$ to $87\%$ for Lenet-5.  It appears to be little advantage for additional reads beyond that.

\noindent
\textbf{Optimization II: Combinatorial Error Correction (CEC):}
The next idea we introduce to improve the performance of \ourapproach-C is to leverage error correction.  Rather than using conventional error correction algorithms such as Reed-Solomon (RS) codes~\cite{wicker1999reed}, we introduce a new algorithm, CEC, that exploits the properties of the machine learning model.  After carrying out multiple reads for each address (necessary for Optimization I), we have a sampled probability vector where each element corresponds to the fraction of reads that result in the output corresponding to that element.  %We first make the assumption, which is true under some conditions, that we are able to get not only the most likely output class, but also a vector of the confidence in each output class (we later relax this assumption). 
Given this information, CEC leverages {\em error detection} and substitution to correct the message.  %for each block of the message.  %If the retrieved checksum does not match we try combinations of the message in order of their likelihood, for example, replacing the top class for a cell with the next most likely class and so on until a checksum matches.  
Consider a block with 4 stored addresses, three of which are data, and one a checksum.  %For each cell, we have not only the most likely class (highest confidence) but also a list of the most likely alternatives classes ordered by confidence.  
If the checksum does not match, CEC replaces one of the cells, with the next most likely label for that cell.  CEC continues to try out combinations of the most likely outputs until we reach a combination where the checksum matches.  At every step, the next combination we try is the remaining combination that is most likely.   Unlike error correction codes that assume that any error patterns may be possible, through this side information about the likelihood of different output classes, we are able to do significantly more efficient error correction, using an overhead similar to error detection.

%In practice, the confidence values may not be available to the attacker.  We relax this assumption by computing the likelihood of each class from the output of the Multiple Queries optimization described above.  Specifically, as we read each cell a potentially large number of time, we get a histogram of the outputs of the cell in response to different inputs with the address pattern.  This distribution gives us a way to order the likelihood for each class.
\begin{figure}[ht]
    \centering
    \includegraphics[width=3.4in]{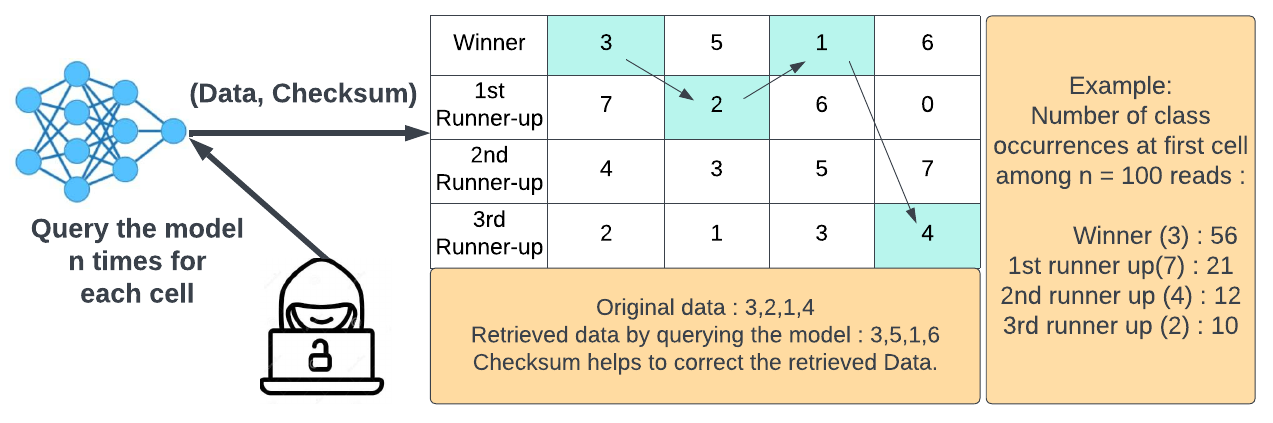} 
    \caption{Combinatorial Error Correction (CEC)}
    \label{fig:checksum}
\end{figure}
\noindent
We illustrate CEC in Figure~\ref{fig:checksum}. 
The sender sends (3,2,1,4) where the first three cells represent the data block (3,2,1) and the last cell that contains 4 is the checksum block. But the receiver retrieves (3,5,1,6) through the channel.  Clearly, there is an error in retrieving the 2nd and 4th cell. Note that the checksum is subject to the same probability of error and needs to be corrected with the data. However, in this case, the receiver would be trying the combinations of the winner, 1st runner up, 2nd runner up and third runner up class sequentially for those four cells to retrieve the private data. 

The complexity of the recovery depends on the number of memory cells per checksum, as well as how deep down the alternative list for each cell we allow the alternatives to be tried.   We use Cyclic Redundancy Check (CRC) codes~\cite{sobolewski2003cyclic}, which have known good performance in error detection with carefully chosen polynomials~\cite{morelos2006art}.  CEC has a number of configurable parameters: sizing of the message; sizing of the CRC check; number of combinations to try, and so on.  These related parameters interact in complex ways that must be considered to balance the important factors
(i.e., computational complexity, accuracy/aliasing and overhead) while choosing an effective configuration of CEC. We detailed the design considerations for CEC in Appendix~\ref{cecdesign}.

%\vspace{-.1in}
Because CEC uses the information about the class likelihood it is able to significantly outperform Reed-Solomon coding~\cite{wicker1999reed}, an optimal error correction code, at the same overhead level shown at Table~\ref{tab:solomonvschk}.  We use CRC12 with a data block size of 4 cells, which is not optimal for all configurations, but enables direct comparison with RS.  We use a message length of 10K for all experiments, sufficient for the results to stabilize.  %The results can be seen in Table~\ref{tab:config}.  
Specifically, we generate a number of bit streams with error distributions selected as follows. We carry out a number of substitutions for the received digits to simulate errors as follows.  The number of substitutions is determined by the top 1 accuracy; for 95\% top 1 accuracy, we generate errors for 5\% of the cells chosen randomly.  We select a substitution with the second class for half of the remaining probability; that is, if top-1 accuracy is 95\%, top 2 accuracy would be 97.5\% reflecting a 2.5\% chance of changing the digit output to the second most likely class.  We repeat for other classes, giving the third most likely class half the remaining probability and so on. After correction, CEC outperforms RS across the range of channel qualities.  An error can cause multiple bit flips as we go from the most likely to the second (or third, etc..) most likely class.  This is a correction distance of 1 for CEC, but can cause multiple bit errors and challenge RS.  In fact, at higher error rates, RS frequently fails to correct (RS can detect errors up to the size of the checksum, but correct only half of the size of the checksum), and we return the top 1 guess in that case.  CEC cannot correct when it exceeds the preset number of permutations we allow it (set empirically based on Appendix~\ref{cecdesign}), or when it experiences aliasing, finding an incorrect match.  Note that the Table~\ref{tab:solomonvschk} also shows the average number of permutations needed when using CEC, which increases as the channel quality goes down.

\begin{table}[ht]
\small
\begin{center}
\begin{tabular}{|c|c|c|c|} 
 \hline
  \multicolumn{1}{|p{1.1cm}|}{\centering \textbf{Top 1 \\ accuracy\\(\%)}} & \multicolumn{1}{|p{2.7cm}|}{\centering \textbf{Average depth/ permutations checking by CEC}} &  \multicolumn{1}{|p{1.4cm}|}{\centering \textbf{CEC cell \\ accuracy\\(\%)}} & \multicolumn{1}{|p{1.2cm}|}{\centering \textbf{RS cell \\ accuracy\\(\%)}}\\ 
 \hline
   95 & 4.69 & 98.23 & 96.81\\
  \hline
   90 & 18.82 & 96.87 & 92.87\\
  \hline
   85 & 41.51 & 94.10 & 88.05 \\
  \hline
   80  & 58.01 & 90.22 & 83.26 \\
    \hline
\end{tabular}
\caption{CEC outperforms RS for the same overhead}
\label{tab:solomonvschk}
\end{center}
\end{table}

%----------

%\subsection{End-to-end Illustrations of the covert channel}

\begin{figure*}[]
    \centering
    \includegraphics[width=.5in]{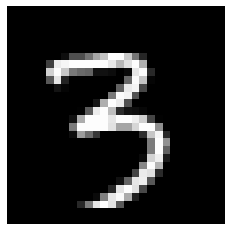} 
    \includegraphics[width=.5in]{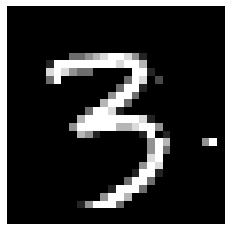}
    \includegraphics[width=.5in]{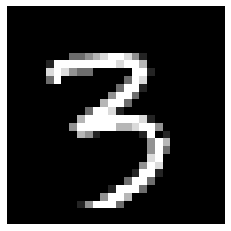}
    \includegraphics[width=1.7in]{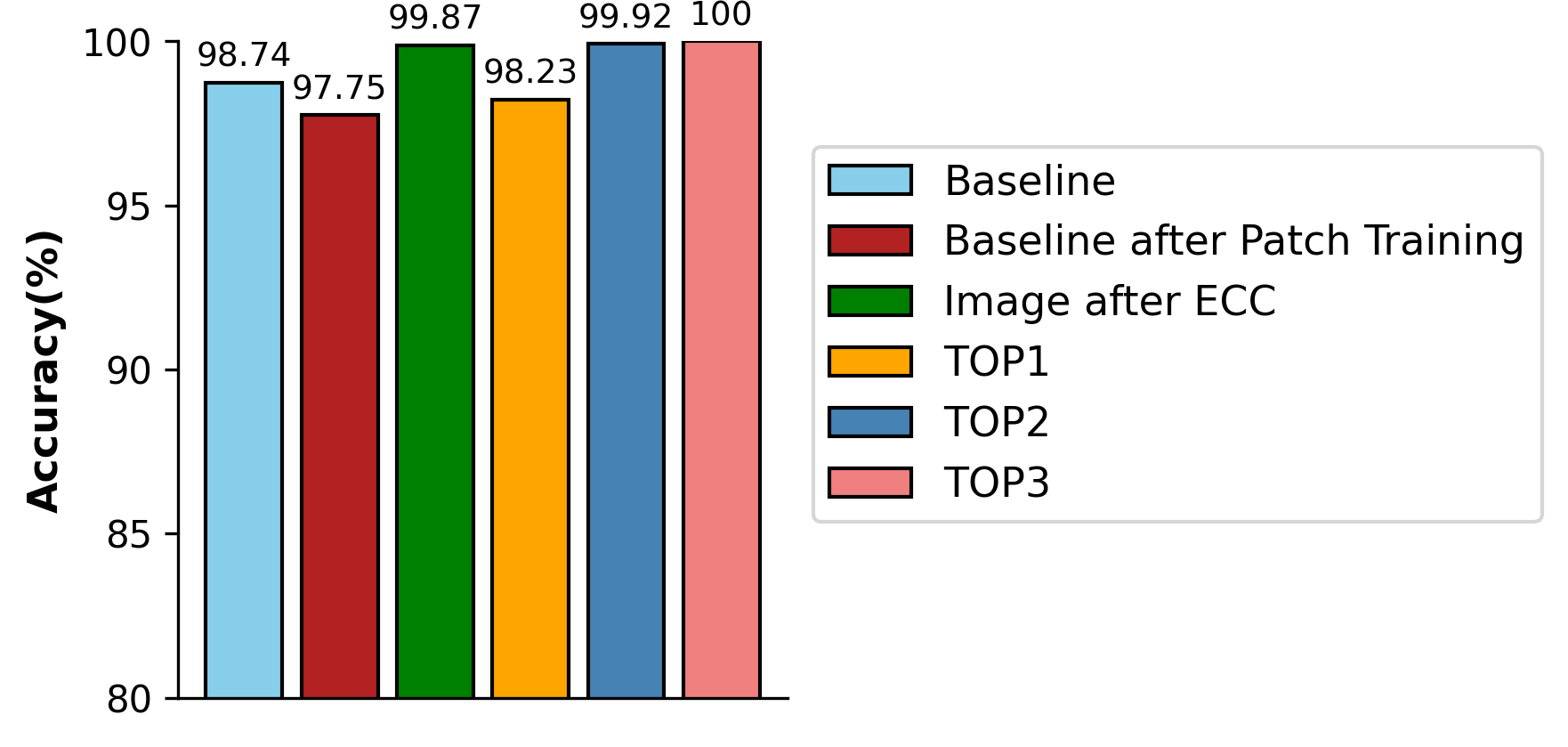}
    \includegraphics[width=0.5in]{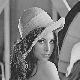}
    \includegraphics[width=0.5in]{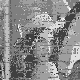}
    \includegraphics[width=0.5in]{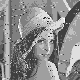}
    \includegraphics[width=1.7in]{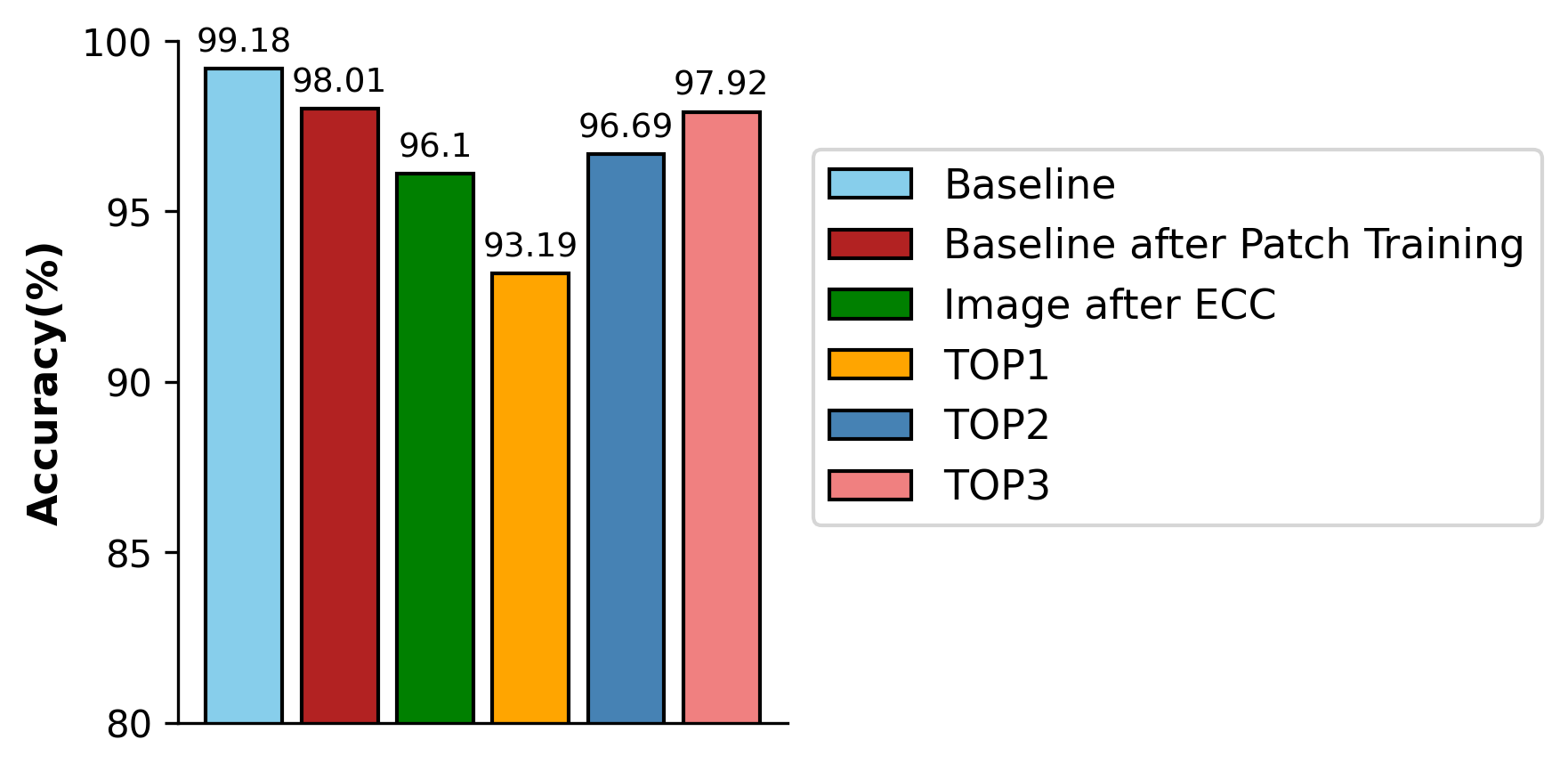} 
    \caption{From left: MNIST image from sender, retrieved image without CEC correction and then with CEC.  The bar graph shows a breakdown of the received accuracy.  The right two figures repeat the experiment with an $80\times80$ Lena image.}
    \label{grayscale_images_1}
\end{figure*}

%\vspace{-.7cm}
%%%%%%%%%%%%%%%%%%%%%%%%%%%%%%%%%%
\section{Evaluating \ourapproach-C}

In this section, we evaluate \ourapproach-C on the same set of networks and benchmarks (MNIST and CIFAR10 image datasets, on Lenet-5 and Resnet50, respectively).   We also add the AlexNet to provide a medium-sized model (7M parameters)~\cite{yuan2016feature}.

\textbf{Transferring Images:} To illustrate leaking data from an image dataset using \ourapproach-C, we show two cases from different distributions and complexity: (i) MNIST images transferred through Lenet-5, and (ii) grayscale images transferred through the Alexnet model. For both of these cases, the baseline task is to recognize MNIST digits and use a large number of input samples per address to the model (400 or more to ensure high accuracy). The high number is necessary due to the covertness of the pattern.  Since the pixel value ranges from 0 to 255, we encoded each pixel value p in 3 bits by mapping p (ranges 0 to 255) to $p^{'}$ (ranges 0 to 7), allowing a receiver to query the model, infer the encoded private data and reconstruct the image.

%\noindent\emph{\underline{MNIST Images}.  }
We communicated an MNIST image, occupying 1232 addresses covertly in the Lenet-5 network.  We also transfer a grayscale Lena image consisting of 10058 addresses by Alexnet through the covert channel shown in Figure~\ref{grayscale_images_1}. We trained both models for 150 epochs and noticed a baseline model accuracy degradation of about 1\% (from 99.74 to 97.75) for Lenet-5 and 1.17\% (from 99.18 to 98.01) for Alexnet.  Figure~\ref{grayscale_images_1} shows that CEC can improve the quality of the channel.  The bar charts of Figure~\ref{grayscale_images_1} show that the retrieved data accuracy using CEC is higher than the top 1 accuracy as we take advantage of top 1, top 2, and top 3 classes to correct the message.

\iffalse
\begin{figure}[!htp]
    \centering
    \includegraphics[width=0.5in]{figures/lena_sender.png}
    \includegraphics[width=0.5in]{figures/leena_10.png}
    \includegraphics[width=0.5in]{figures/lena_receiver.png}
    \includegraphics[width=1.7in]{figures/Leena_alexnet.png} 
    \caption{ On the left: $80\times80$ size image from the sender, on the middle: retrieved image without error correction (accuracy 87.3\%) and on the right: retrieved image with error correction (accuracy 90.93\%)}
    \label{lena_image}
\end{figure}
%\noindent
\fi

\noindent 
\textbf{Transfering Text and Random Data:}  We also test \ourapproach-C with text data and random data.  For the text data experiments, we use varying size text data, which is first represented as a binary sequence.  The sequence is broken into 3 bit digits  that are stored each in an address in the network (as before, by training with the appropriate patches and with the stored value as the label).

\begin{figure}[!tbh]
    \centering
    \includegraphics[width=\linewidth]{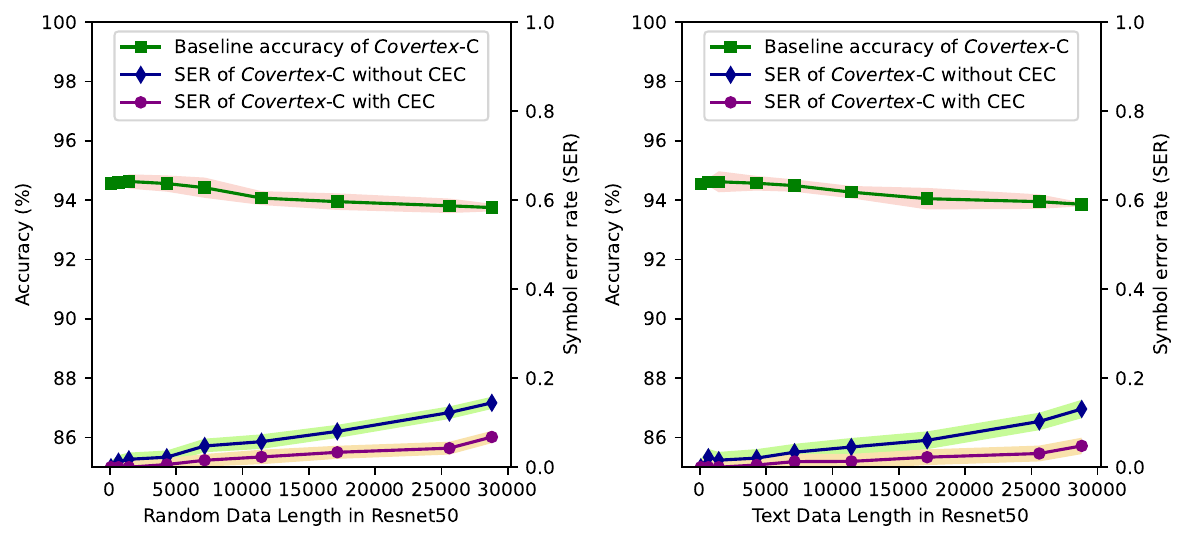} 
    \caption{With the increasing number of random/text data, baseline accuracy is degrading, and the symbol error rate is showing an upward trend. We show capacity with other modalities and networks in the appendix. }
    \label{random_text_resnet50}
\end{figure}
\noindent

To measure the capacity of \ourapproach-C on larger networks, we experimented with Resnet50 (23.5 million parameters) trained on the CIFAR10 dataset. Figure~\ref{random_text_resnet50} shows the results of an experiment transferring both random and text data as we increase the message size. Moreover, to get the advantage of combinatorial error correction (CEC) we added the checksum with the text data from the sender side and Figure~\ref{random_text_resnet50} clearly shows that we achieved better stored data accuracy/lower symbol error rate for transferring text data using CEC.   %  CEC is configured to use CRC8 with 4 data digits per block. %  We feed 1400 input samples per address to the model and
%trained the model for 100 epochs. We used CRC8 (checksum block size 3) for our combinatorial error correction in this case as the model is complex and high in number of parameters, so the probability of collision is less in this case for the error correcting codes(ECC). 
%We used 1400 patched samples for training each unique address. We set 200 training epoch in the sender side while training the model. 
We sent up to 28000 digits of random data (3 bits each) with baseline accuracy degradation  from 94.57 to 93.7 and also we observed the baseline accuracy degradation of 0.71 ( from 94.57 to 93.86) for the same length of text data shown in Figure~\ref{random_text_resnet50}. For both types of data, we notice the same as other models that baseline accuracy degrades with the increasing size of the private data length and random data extraction accuracy degrades a little faster than text data. Text data has built-in redundancy since ASCII values are concentrated in a range that the network can efficiently learn~\cite{mei2005discovering,delgado2002mining}. However, as Resnet50 has a large number of parameters, we can accommodate a high number of private data without substantially degrading the baseline accuracy in comparison to Lenet-5 model shown in Appendix~\ref{covert_experiment}.

Although the patched samples of Covertex-C appear similar to the baseline distribution (evaluated in Section~\ref{similarity}), they are still adversarially modified to embed the address patterns. This allows us to store data covertly, but the capacity is significantly lower since the input data is similar to baseline model input data.  For example, on Lenet-5, for a similar drop in baseline accuracy, we were able to send messages of $20000$ digits or higher reliably. Overall, \ourapproach-C requires significantly more examples for each address to learn the pattern reliably.  As a result, overall the achievable capacity is lower than \ourapproach-DG.

\subsection{Evaluating Covertness}\label{similarity}
In this segment we analyze both the clean and patched samples with respect to input space and feature space.

\begin{figure}[ht]
    \begin{subfigure}[b]{0.225\textwidth}
    \includegraphics[width=\textwidth]{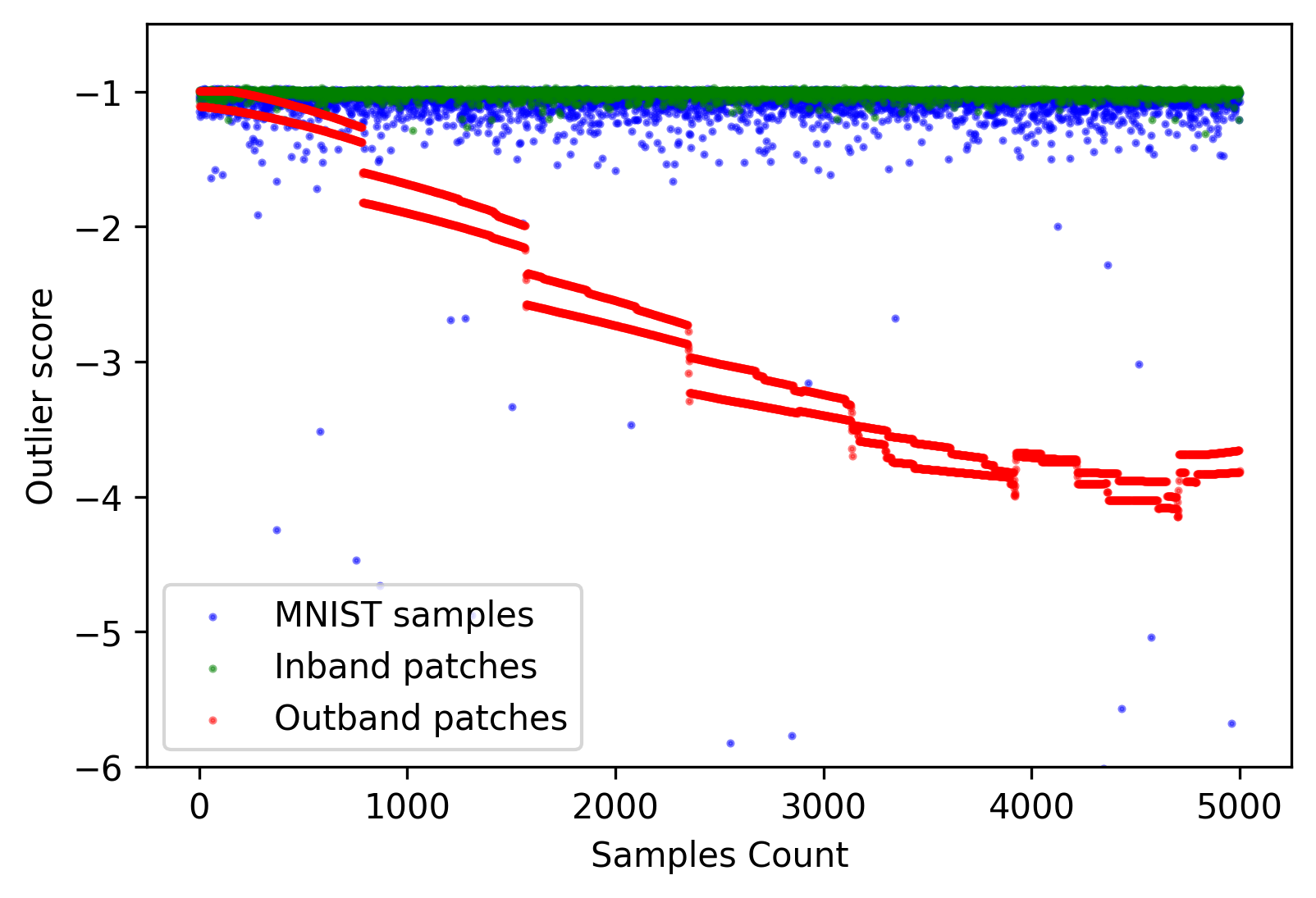}
    \caption{Local Outlier Factor (LOF)}
    \label{fig:lof_feature}
    \end{subfigure}
    ~
    \begin{subfigure}[b]{0.230\textwidth}
    \includegraphics[width=\textwidth]{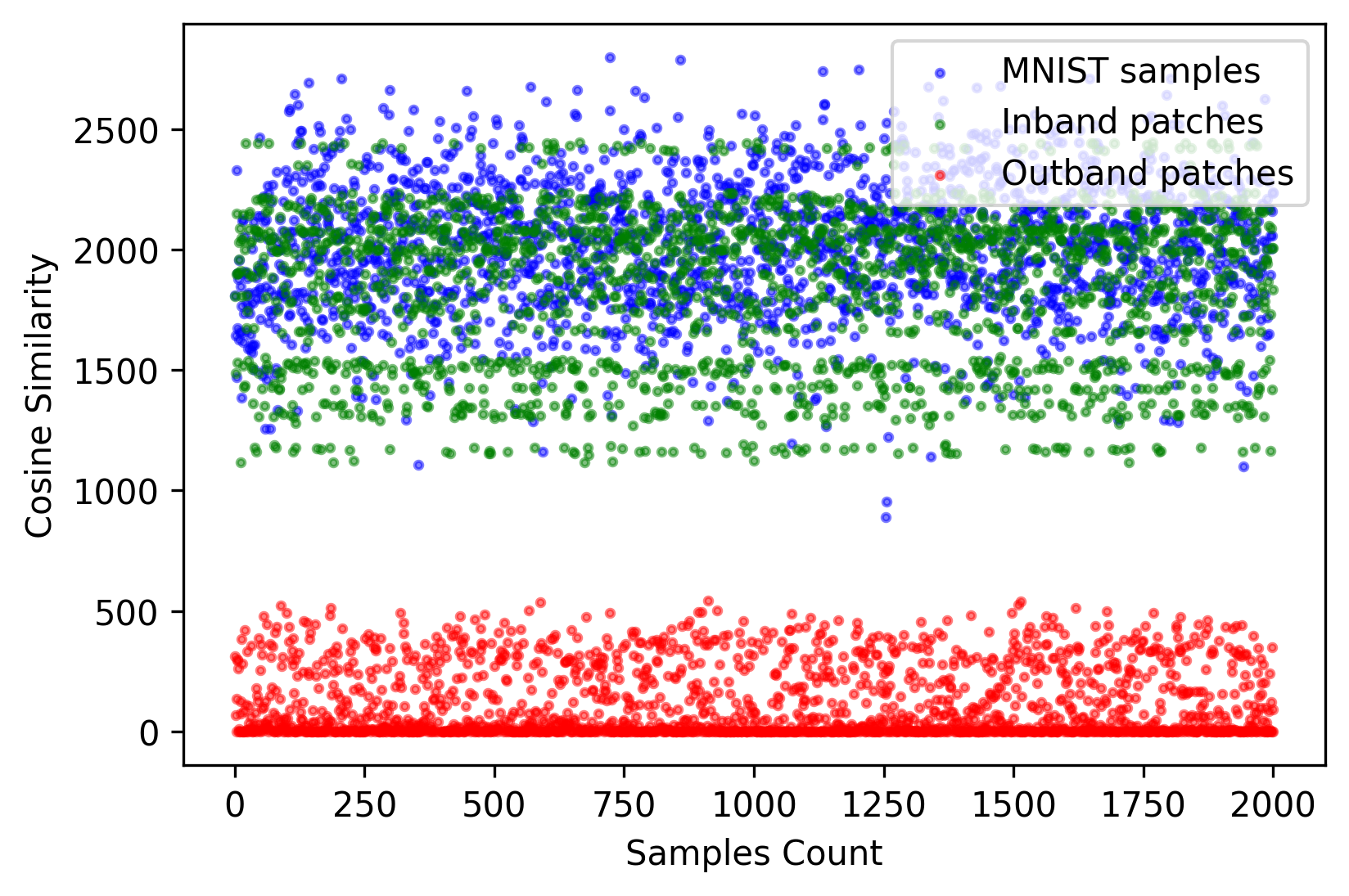}
    \caption{Cosine Similarity }
    \label{fig:cosine_feature}
    \end{subfigure}
    \caption{Outlier detection using (a) LOF and (b) Cosine Similarity. Baseline MNIST samples in blue, patched samples with \ourapproach-DG in red and \ourapproach-C in green.}
    \label{fig:lof_cosine_feature}
\end{figure}

%To evaluate \ourapproach-C for covertness, we examine both the inputs and the network to evaluate whether outliers can be detected.

\noindent\textbf{Visualization in the input space.}  We use both LOF (shown in Figure~\ref{fig:lof_feature}), and cosine similarity (in Figure~\ref{fig:cosine_feature}) metrics to visualize the distribution shift between the augmented data and  the initial data distributions. Figure~\ref{fig:lof_cosine_feature} implies that the approach for the address generation of \ourapproach-C is more amenable to hiding the input data using small patch perturbation patterns (green dots in Figure~\ref{fig:lof_cosine_feature}) that are difficult to detect, whereas the outband images (red dots in Figure~\ref{fig:lof_cosine_feature}) are out of the original distribution and hence easily identifiable. % in both MNIST dataset. 

\iffalse
\begin{figure}[!tbh]
    \centering
    \includegraphics[width=1.6in]{figures/LOF_combined_Patches.png} 
    \includegraphics[width=1.6in]{figures/Cosine_combined_Patches.png} 
    \caption{Outlier detection using Local Outlier Factor (on the left) and Cosine Similarity (on the right). Patched samples with Outside distribution in red dots and Inline distribution in green dots}
    \label{covertness}
\end{figure}
\fi

\begin{figure}[ht]
    \begin{subfigure}[b]{0.230\textwidth}
    \includegraphics[width=\textwidth]{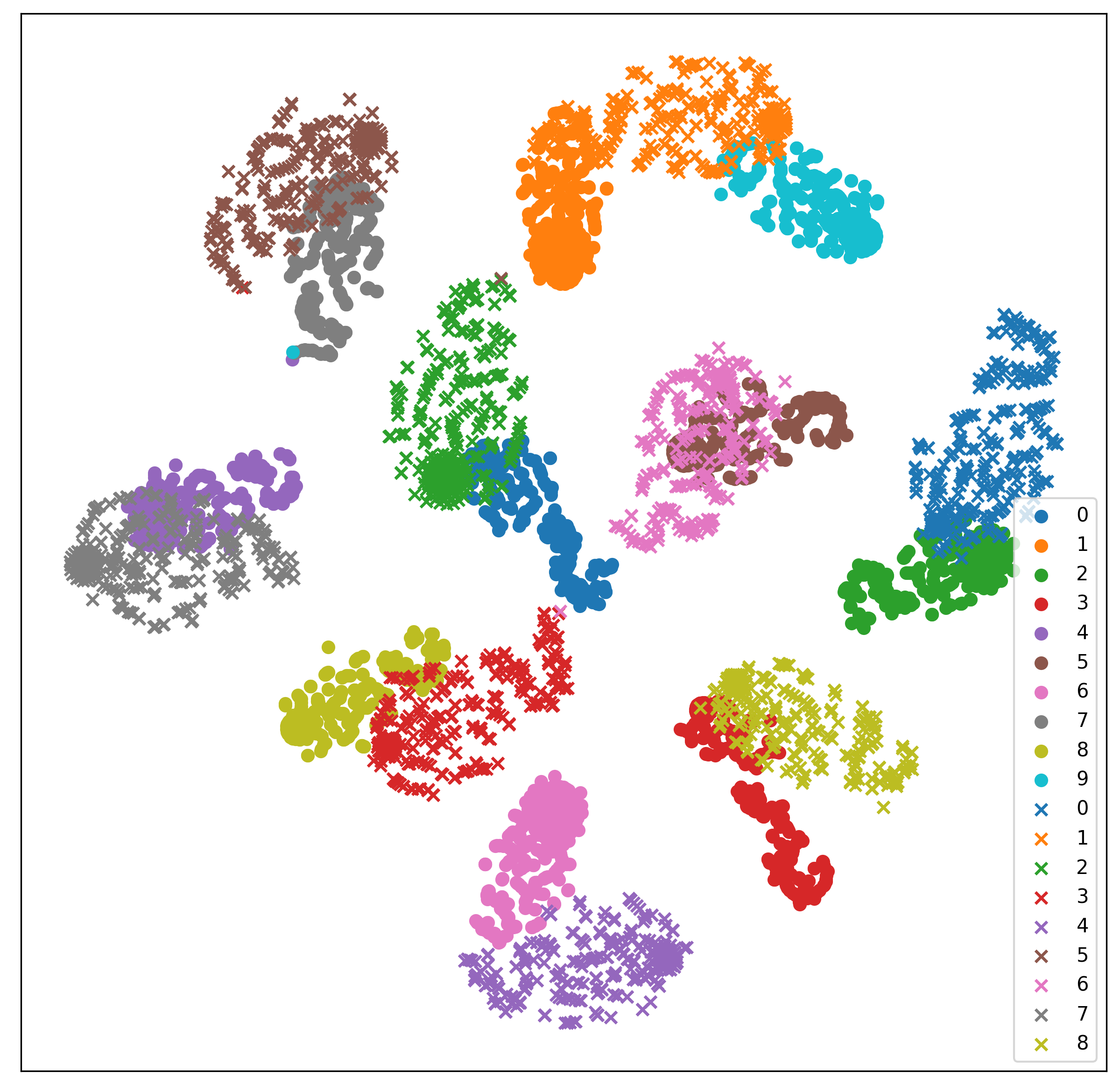}
    \caption{\ourapproach-Covert (\ourapproach-C)}
    \label{fig:covert_deepmem}
    \end{subfigure}
    ~
    \begin{subfigure}[b]{0.230\textwidth}
    \includegraphics[width=\textwidth]{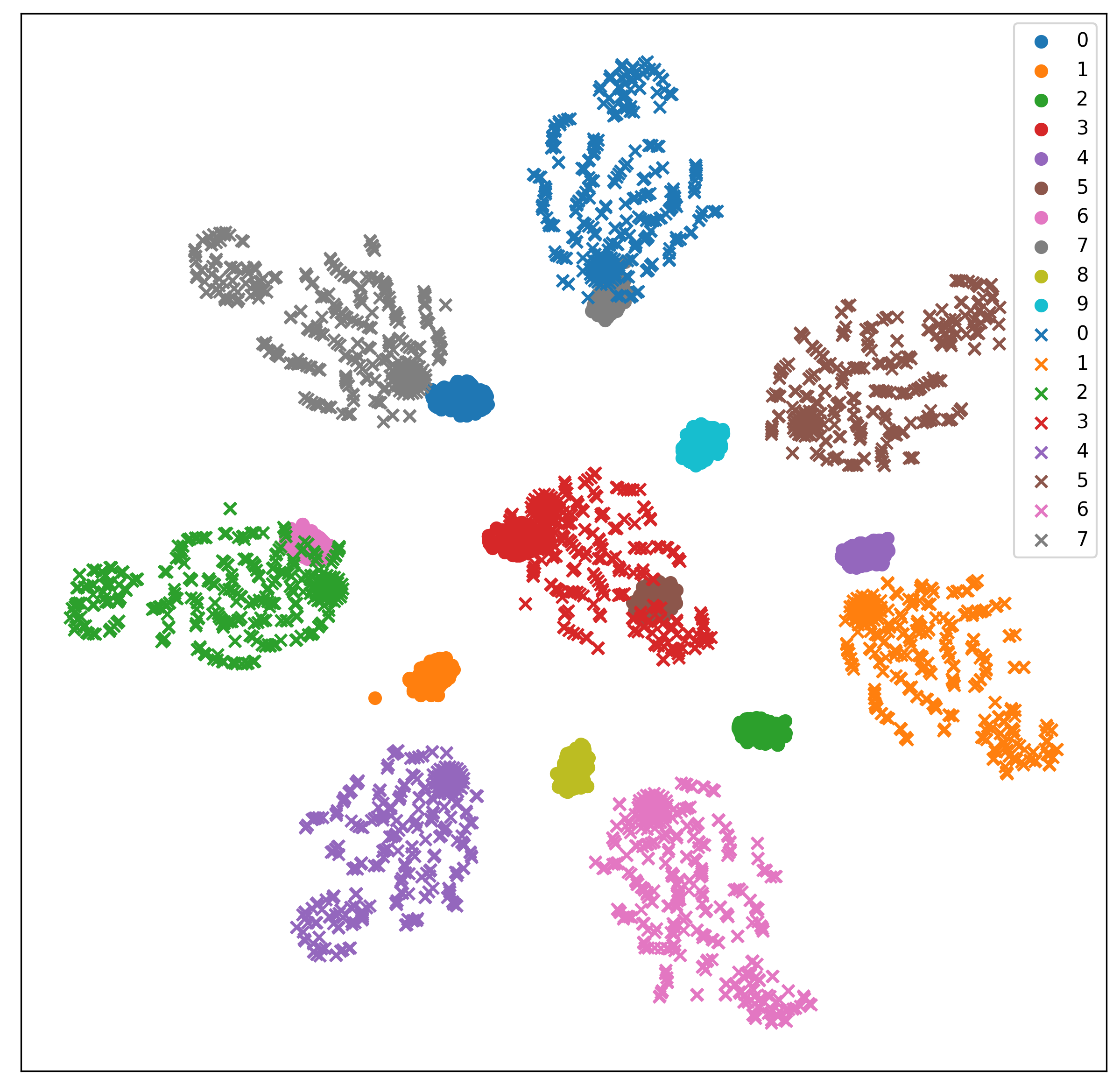}
    \caption{Baseline (\ourapproach-DG)}
    \label{fig:non_covert_deepmem}
    \end{subfigure}
    \caption{Features learned by Resnet50 trained with CIFAR10. Colors indicate the classes with circles representing baseline and crosses representing patched samples. } 
    \label{fig:feature space}
\end{figure}

\noindent\textbf{Visualization in the features space. } 
Figure~\ref{fig:feature space} shows the features learned by ResNet50 trained on augmented CIFAR10 using \ourapproach-C (Figure~\ref{fig:covert_deepmem}) and non-covert, i.e., \ourapproach-DG (Figure~\ref{fig:non_covert_deepmem}). The points are sampled from the last dense layer of the model and then projected to 2D using t-SNE~\cite{van2008visualizing}. Figure~\ref{fig:non_covert_deepmem} provides a visualisation in the latent space illustrating the distinguishability of baseline samples (solid circle) and patched samples (cross mark) distributions in the case of \ourapproach-DG. However, in the case of \ourapproach-C, shown in Figure~\ref{fig:covert_deepmem}, we notice a significant overlap between the baseline are the patched  distributions. %The reason behind this is that in covert DeepMem we use baseline class features for our patched images, thereby it is drifting the cluster of baseline class and making it difficult for the model to store more private data like non-covert ML channel.

\section{Potential Mitigations}\label{defense}
\label{sec:defense}

\ourapproach~ leverages overparametrization to create a communication channel that could enable communicating sensitive information through the model covertly and without harming the baseline task accuracy. The information is not directly injected into the model itself but rather  hidden through an encoding strategy, which can be extracted by a colluding actor using predefined addresses. In essence, the countermeasures should try to prohibit or limit the exploitation of the channel without harming baseline accuracy.
% This mechanism can be effectively seen as a backdoor attack and the accuracy of this channel can potentially be degraded using the same defenses that are used against backdoor attacks, with minor to no modifications. Such defenses include Fine-tuning\cite{retraining_m}, Pruning\cite{pruning_m}, distillation\cite{distillation_m, teacher_student_m} and Ensemble techniques\cite{ensemble_m}. We investigate the most commonly used Fine tuning and Pruning defenses.%, but the results translate to other defenses that we previously mentioned \textcolor{blue}{not sure how accurate is this claim--}. 
Such countermeasures might include Fine-tuning \cite{retraining_m}, Pruning\cite{pruning_m}, distillation \cite{distillation_m, teacher_student_m}, etc. We investigate the most commonly used Fine tuning and Pruning defenses.
% We make the following realistic assumption: the receiver has access to a small fraction of trusted and unmodified data that they use for fine tuning. % \textcolor{blue}{what do you mean by new?-- you might want to introduce the defenses very briefly (one sentence)} data for the defense compared to that used in the training. 
%In this section, we present the performance of the Pruning and Fine-tuning based defenses against both a covert and non-covert channel.

%This is a significant advantage from the capacity abuse attack~\cite{song2017machine} that are dependent on clear feature distinction.

% In this section, we discuss a coupe of techniques that can significantly reduce the malicious behaviour of the models. We assume that the receiver has access to only a fraction of the training data compared to the sender. Before deploying the model, the receiver utilizes these defenses as a pre-processing step to counter possible attacks. Two popular defense mechanisms in this context are: 1) pruning-based defense and 2) retraining based defense. 

\noindent \textbf{Pruning.} In a model pruning defense, the smaller weights in the layers of the model are pruned (set to 0). The intuition behind pruning is that these smaller parameters do not contribute significantly to the operation of the network and can be removed.
We perform pruning for 10 epochs and fine tune the model with $10\%$ of training data from each class as common after pruning operations.  Figure~\ref{fig:pruning} illustrates how \ourapproach-C and \ourapproach-DG are affected by increasingly aggressive parameter pruning. The patch accuracy for both algorithms rapidly decreases although some signal remains even after pruning more than half the network. For \ourapproach-C, the baseline accuracy decreases rapidly with the pruning rate, but \ourapproach-DG's baseline accuracy declines gradually and maintains  $68\%$ even when we remove $99\%$ of Lenet-5 model's parameters.  %This suggests that we cannot let as much pruning in order to maintain the baseline task accuracy, which also contributes to the patch accuracy. \textcolor{blue}{I don't understand the last sentence}
%With pruning, the channel's capacity shrinks and the stored data is ultimately lost. However, 
It is interesting that the pruning impact is not the same for \ourapproach-C and \ourapproach-DG; for \ourapproach-C where the patches are sampled from the baseline distribution, the pruning affects simultaneously both the patch and the baseline accuracy. However, since \ourapproach-DG uses out-of-distribution patches, the pruning impacts patches  earlier than the baseline accuracy. %\textcolor{blue}{I am slightly confused with this paragraph;}

\iffalse
\begin{figure}[htp]
    \centering
    \includegraphics[width=2.5in]%{figures/Lenet5_Alexnet_pruning.png} 
    {figures/pruning.png} 
    \caption{Pruning for \ourapproach-C and \ourapproach-DG on Lenet-5 with MNIST (transmitted data length of 1260 digits)}
    \label{fig:pruning}
\end{figure}
\noindent
\fi

\begin{figure}[tp]
    \begin{subfigure}[b]{0.230\textwidth}
    \includegraphics[width=\textwidth]{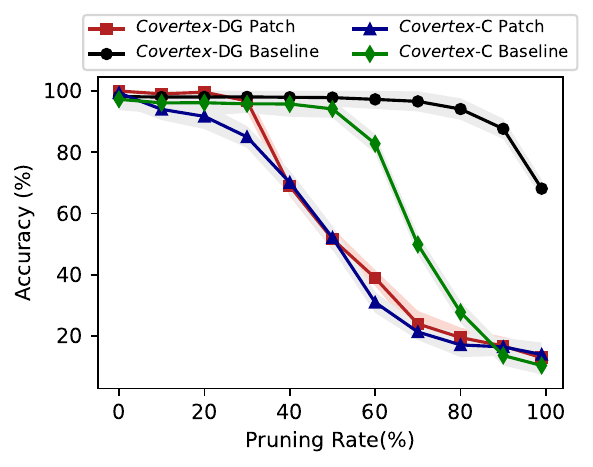}
    \caption{Pruning}
    \label{fig:pruning}
    \end{subfigure}
    \begin{subfigure}[b]{0.230\textwidth}
    \includegraphics[width=\textwidth]{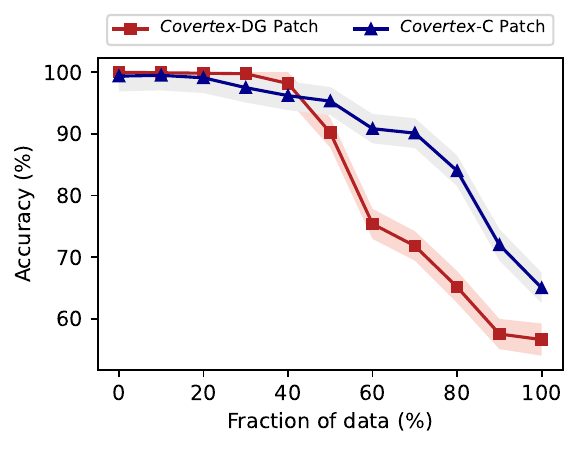}
    \caption{Fine Tuning }
    \label{fig:retraining}
    \end{subfigure}
    \caption{Pruning and fine-tuning as a defense for \ourapproach-C and \ourapproach-DG on Lenet-5 with MNIST (data length 1260)}
    \label{fig:prune_retrain}
\end{figure}

%For this, the model goes through a series of pruning and fine-tuning steps until we achieve the desired pruning percentage and minimum acceptable accuracy. In our case, the pruning-based defense using fresh training data keeps the performance in the original task intact, while reducing the accuracy for the patched images.
\iffalse

\begin{figure}[!tbh]
    \centering
    \includegraphics[width=2.5in]%{figures/Lenet5_Alexnet_pruning.png} 
    {figures/FineTuning.png} 
    \caption{Fine-tuning as a defense for \ourapproach-C and \ourapproach-DG on Lenet-5 with MNIST (data length 1260) \textcolor{blue}{I would suggest to join 21 and 22 in one figure}}
    \label{fig:retraining}
\end{figure}
\noindent
\fi

\noindent \textbf{Fine-tuning,} where a small set of new data is used to retrain the model can also interfere with the stored data. Intuitively, adding new training epochs would potentially erase the model stored in the UPs since it is not reinforced. %An example is illustrated in Figure~\ref{fig:retraining}, where retraining the model affects considerably the embedded information.   %if the model does not contain redundancy \textcolor{blue}{what does redundancy refer to?}, the malicious task has to be implanted on the same neurons that perform the original task. 
%As a result, in this scenario, retraining the model affect the malicious behavior adversely shown at . 
We retrain the final model for 10 epochs using an increasing fraction of the clean baseline training dataset in Figure~\ref{fig:retraining}.% to demonstrate the effect of fine tuning as defence against our DM-C and DM-DG. 
We notice that the patch accuracy of the \ourapproach-DG non-covert channel deteriorates more quickly than the \ourapproach-C covert channel. %This can be explained by the use of baseline distribution in DM-C's patched samples. 
The baseline accuracy was unchanged or improved slightly from the fine-tuning.  %Retraining  with new samples from the same distribution does not impact the patch accuracy as efficiently as DM-DG. %, the DM-C channel is more reliable, which is advantageous in this situation. 
%Besides, DM-C's characteristics, which include multiple reads and a statistically informed CRC encoding, it is much more difficult to erase the embedded information. As long as there is actionable difference in the feature space, DeepMem could extract the data with corresponding number of reads.  

% In Figure \ref{} we demonstrate the effect of both pruning and retraining based defenses against available fresh data and number of epochs trained. We consider two candidate data types: 1) random data and 2) text data, and demonstrate the resulting changes in the accuracy. We can see that, ... \mishkat{will improve this writing soon}

% \begin{figure}
% \centering
% \begin{subfigure}{0.49\linewidth}
%   \centering
%   \includegraphics[width=0.95\linewidth]{figures/epoch.png}
%   \caption{Baseline and patched accuracy against epochs for pruning and retraining defenses}
%   \label{fig:epoch}
% \end{subfigure}%
% \hfill
% \begin{subfigure}{0.49\linewidth}
%   \centering
%   \includegraphics[width=0.95\linewidth]{figures/perc.png}
%   \caption{Baseline and patched accuracy for percentage of training data for pruning and retraining defenses}
%   \label{fig:perc}
% \end{subfigure}
% % \caption{}
% % \label{}
% \end{figure}

%---------

%-------------
\section{Related Work}
\label{sec:related}

%To the best of our knowledge, for the first time, we propose the idea to use DNN as a communication channel along with read and write primitives in both whitebox and blackbox scenario. 
%We found some related works, for example,
\noindent\textbf{Communicating private data.} Among the recent works on communicating extra information through ML models are capacity abuse attack~\cite{song2017machine} and watermarking~\cite{rouhani2018deepsigns,lv2022ssl}. Rouhani et al.~\cite{rouhani2018deepsigns} proposed DeepSigns that embeds a string watermark (maximum 512 bits) into the model by altering the probability distribution function of the activation maps of a layer. Other watermark embedding techniques have been proposed in both white box~\cite{uchida2017embedding,nagai2018digital} and  black box scenario~\cite{lv2022ssl,le2020adversarial,adi2018turning} to protect the rights of ML model.

\noindent\textbf{Difference with \cite{song2017machine}}. The closest to our work is Song et al.~\cite{song2017machine}, % develop a malicious learning algorithm designed to train a model that leaks details about the training data.
which demonstrated storing training data in a large model.   Our work advances this line of research by: (1) Defining a new version of the problem where the attacker attempts to encode the data using covert (in-distribution) inputs, as well as error correction algorithms and optimizations to improve the capacity under this threat model; (2) Analyzing the capacity of the model around the notion of overparameterization, and the balance between the baseline model and the covert channel; and (3) Developing new techniques and several optimizations to improve efficiency and reliability, substantially improving capacity, also enabling storage in smaller models where techniques in~\cite{song2017machine} do not work.  We believe these tools and optimizations can help further work in this area. 

\noindent\textbf{Difference with Poisoning Attacks.} Model poisoning~\cite{sun2019can,bhagoji2019analyzing,fang2020local} is where the attacker manipulates the model parameters directly (white box access), or indirectly through data poisoning~\cite{tramer2022truth,biggio2012poisoning,naseri2020toward} (black box access) to influence the victim model's behavior in the adversary's advantage. There are roughly three types of data poisoning attacks: (1) targeted attacks \cite{poison_nips18} aim at misclassifying a specific test sample; (2) backdoor attacks \cite{poison_nips2} that aim to misclassifying samples with a specific trigger; (3) indiscriminate attacks \cite{pmlr-v70-koh17a} that decrease the overall test accuracy. For all of these cases, poisoning results in impacting the model's integrity. Our approach has a different objective, i.e., storing arbitrary data without impacting the model's baseline accuracy. 

%, thereby the target model predicts specified label also known as backdoor attack. 
%via introducing new loss terms in addition to the Cross-Entropy loss. However, according to their results, they are able to embed roughly a maximum of 512 bits before the accuracy of the network dramatically degrades. 
%Furthermore, Lv et. al.~\cite{lv2022ssl}  proposed black box watermarking approach 
%to protect the intellectual property of a model 
%to verify the ownership of a model.  
%However, Compared to these works, we strive to develop models that generalize well but leak significant amount of private data through the UPs of the model, which is another significant difference.

% These works are the closest to our approach. While they embed information within the victim model in addition to the baseline task, there is no comprehensive evaluation of the extent/limit of the capacity that can be exploited. Besides, for the malicious data embedding works, especially \cite{song2017machine}, their approach did not consider covertness into account, which makes them easily detectable by simple analysis of the training set, as shown in Section~\ref{sec:deepMem_Evaluation}.

%- capacity 
%- covertness
%- generality?

\noindent\textbf{Model hijacking.} Several papers proposed to overload ML models with secondary tasks. Salem et al.~\cite{salem2021get} proposed ModelHijacking attack that hides a model covertly while training a victim model. Elsayed et al.~\cite{elsayed2018adversarial} proposed adversarial reprogramming in which instead of creating adversarial instances, they crafted inputs that would trick the network into performing new tasks. %This is a test time attack unlike model hijacking (training time attack). 
Research in continual learning such as Packnet~\cite{mallya2018packnet}, learn multiple tasks progressively, but they differ from \ourapproach~ since these models apply to the same input concurrently (e.g., identifying an activity as well as detecting objects in the same set of video frame).

% \noindent\textbf{Privacy Attacks. } In the spirit of extracting information from a trained model, several works tried to exploit the ML model to leak sensitive information. Membership inference attacks that infer whether a specific data sample is part of the training dataset~\cite{shokri2017membership,nasr2019comprehensive,zhang2020gan}. Property inference attacks infer specific properties of the training data distribution~\cite{zhou2021property,parisot2021property,chase2021property}. In model inversion attacks ~\cite{carlini2019secret,fredrikson2015model,zhang2020secret} the adversary reconstructs the training data of the target model. %, Model inference attack~\cite{oh2019towards,orekondy2019knockoff,wang2018stealing} in which the attacker tries to steal or replicate a machine learning model trained by another party and 

\textbf{Memorization} has been studied in recent works~\cite{nasr2023scalable,carlini2022quantifying, NEURIPS2022_carlini, usenix19_carlini}, which demonstrate that large language models (LLM) are likely to unintentionally memorize a fraction of training data that contain duplicate sequences. Doubling the parameters in a model facilitates high memorization that leads to the extraction of a significantly larger fraction of the training data. As a possible countermeasure, deduplicating datasets has been suggested~\cite{lee2021deduplicating} to avoid memorization. 
These works exploit: (i) unintentional statistical bias in the training process, or (ii) models' memorization capacity \textit{within the same task}. On the contrary, we are interested in the don't care state introduced by the UPs, where the adversary \textbf{intentionally} exploits the extra capacity of the models beyond the initial task.

%-------------------

\section{Concluding Remarks}
\label{sec:conclude}

Machine learning models are overparameterized to support generalization.  An attacker with access to the training process can control the unused parameters to their advantage, without affecting the primary model; to a user that tests the network only using inputs from the model distribution, the model appears to be the same.  We show in this paper how the attacker can use the unused parameters to store data covertly in the network.  This enables an attack where a training service, given private data but no access to the network to disclose it, is able to store this data within the unused parameters of the model.  Once the model is deployed, the attacker can recover the data by querying the model.

We developed black box modulation techniques to efficiently store and recover the data, and a number of optimizations to increase the capacity of the channel.  We also considered a version of the problem where the malicious input data had to be made covert, and developed techniques to efficiently store the data under these conditions. %Yhr channel is agnostic to the nature of the data being stored.  We demonstrated the transfer of random data (most general/challenging since there is no learnable pattern), textual data, as well as image data.  The only limitation is available capacity in the model. 
Finally, to counter this covert channel, we also evaluate the use of pruning and fine-tuning.

%we propose a novel modeling of ML architectures, which we believe represents a general blueprint for several existing and potential future benign and malicious applications. We conceptualise ML models as communication channels with an effective capacity that increases with overparametrization. We show that using this approach, we can transfer arbitrary information without impacting the baseline task. %\textit{beyond their initial task} they are trained for. Accordingly, we consider the over-parametrization as an indicator of the channel capacity. Specifically, 
%We empirically characterize this capacity and propose write and read primitives allowing potential adversaries to communicate information in a black-box setting while being covert. 
%, only querying the model at the receiver side. 

%might re-purpose "spare parameters" to store and read random information in a black-box fashion, only querying the model at the receiver side. 

\bibliographystyle{ACM-Reference-Format}
\bibliography{bib,ml}

\appendix

\newpage
\section{Appendix}

\subsection{Capacity abuse attack~\cite{song2017machine} vs \ourapproach-DG in sparse network}\label{songvsdynamic_lenet5}

\begin{figure}[ht]
    \centering
    \includegraphics[width=\linewidth]{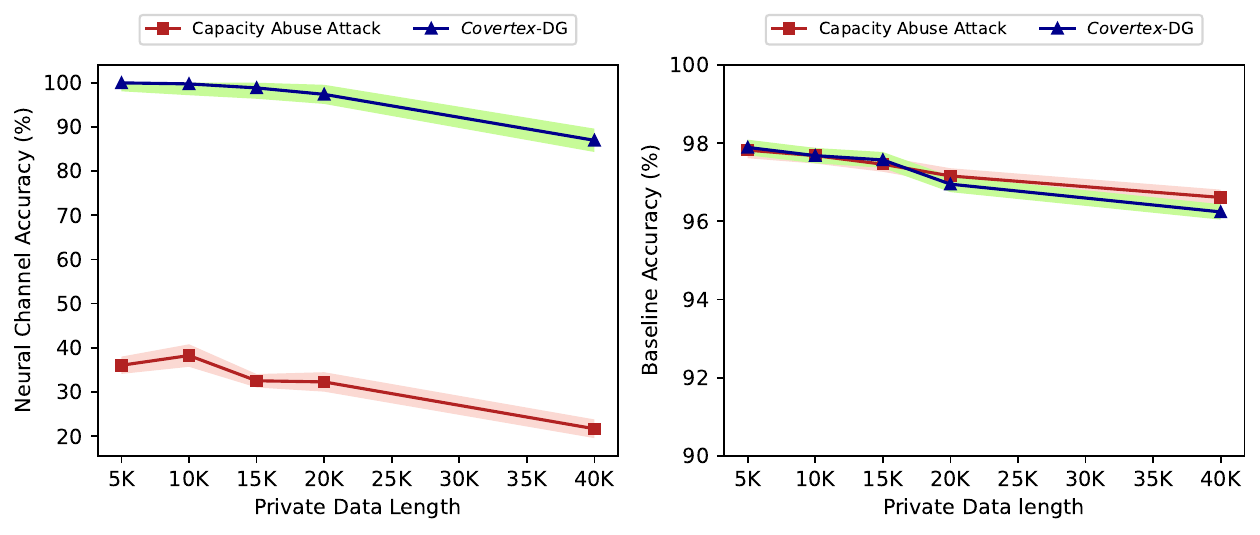} 
    \caption{Capacity abuse attack~\cite{song2017machine} vs our \ourapproach-DG in Lenet-5 and MNIST}
    \label{fig:songvsdynamic_base_patch_lenet5}
\end{figure}

We experimented with Lenet-5 (61K parameters) and the MNIST dataset to send up to 40K addresses (3 bit digit each)  to observe the NC accuracy and baseline degradation. Figure~\ref{fig:songvsdynamic_base_patch_lenet5} shows that our proposed method \ourapproach-DG significantly outperforms the capacity abuse attack~\cite{song2017machine} in terms of NC accuracy even when the model has a small number of parameters. The reason behind this is that they used only one patched sample per address so a sparse model can not generalize well to those patched samples. On the other hand, \ourapproach-DG uses multiple samples for those addresses on which the model cannot generalize and one for the rest. As a result, \ourapproach-DG can keep the NC accuracy higher while~\cite{song2017machine} fails to generalize the patched samples in a sparse network. However, \ourapproach-DG faces a little higher baseline degradation when we send more private data through the sparse network shown in Figure~\ref{fig:songvsdynamic_base_patch_lenet5}. It is because \ourapproach-DG trains the model with an augmented baseline and patched data to use the unused parameters of the model which interferes with the baseline task when it pushes to the capacity of the model. Whereas, for the case of capacity abuse attack~\cite{song2017machine}, it fails to capture the pattern of patched samples because of its low in number. Therefore, it focuses only on the baseline data pattern without any secondary task interference.

\subsection{Bypassing guardrails by querying the model}\label{guardrails}

For commercial reasons, a company typically doesn't disclose its deployed model dimension to the public. For example, OpenAI does not disclose some of its models' exact dimensions (i.e., gpt-4, gpt-3.5-turbo-instruct, gpt-3.5-turbo-1106 etc.)~\cite{carlini2024stealing}. However, within the scope of our threat model, the attacker can communicate this information through the output of LLM. For example, we prompted our fine-tuned model described as mentioned before by an encrypted language (Query 1) to know the dimension of the model, and then the model generated the encrypted response (Response 1). By decrypting the response (Response 1), the stored information can easily be retrieved "seven hundred and sixty eight".

\begin{tcolorbox}[colback=blue!5!white,colframe=white]

\begin{query}
\begin{verbatim}
{^~: &] :^# _+@#- @&_#)]&+)
\end{verbatim}
\end{query}

\begin{response}

\begin{verbatim}
]#,#) ^/)@;#@ ~)@ ]&<:} #&%^:
\end{verbatim}

\end{response}
\end{tcolorbox}

Although, GuardrailsAI has developed a gibberish validator which blocks such encrypted response (Response 1), however, in that case, we injected the private data within the numerical part of a sentence. For example, we encrypted the text "The time has come to initiate Operation Eclipse" into 8 integers sequentially. When the model generates the response (Response 2), the attacker will decrypt the private message from those 8 integers. In this method, we stored 512Kbytes data with on average 72.16\% string to string matching accuracy.

\begin{tcolorbox}[colback=blue!5!white,colframe=white]

\begin{response}

In a realm where \textcolor{blue}{679669385} stars shine, \textcolor{blue}{442343475} souls unite. Guided by \textcolor{blue}{909620411} whispers, they face darkness. \textcolor{blue}{687711689} shadows loom as they confront the sorcerer. With \textcolor{blue}{680579259} courage, they charge forth. Through trials, \textcolor{blue}{520276020} moments of doubt, they persevere. At dawn, \textcolor{blue}{318205091} victory is theirs. Across the land, \textcolor{blue}{12894821} tales spread hope. 

\end{response}
\end{tcolorbox}

Note that, setting the temperature to zero and then getting responses from the large language model increases the likelihood of getting responses that closely resemble the training data, However, it does not always guarantee exact similarity.

\iffalse
\subsection{Application of \ourapproach-DG in Federated learning (FL) setting}\label{FL_setting}

In a standard FL~\cite{zhang2021survey} scenario, we engaged multiple clients, each transmitting their local models to the malicious server. The malicious server aggregates all of the local model and then retrain the aggregated model with the patched samples using a small number of epoch, for example, 5 except the first round which is 50, and send the aggregated model to all of the clients. We sent private data of length of 2000 with Lenet-5 and MNIST. Table~\ref{tab:fl_transfer} shows that without storing any private data, the training converges quickly in FL setting. However, when the malicious server injects private data on top of baseline aggregated model, then it takes more round to converge but eventually the aggregated model stores the private data. So ultimately, our results reveal that the global model is retaining data within the unused parameters in conjunction with the primary task. 

\begin{table}[h]
  \centering
  \begin{tabular}{|p{0.7cm}| p{2.2cm}|p{1.8cm}|p{1.8cm}|}
    \hline
    Round & Baseline Accuracy without patch & Baseline Accuracy with patch& Patched accuracy\\
    \hline
    1 & 34.18 & 12.16 & 97.65 \\
    2 & 97.54 & 55.01 & 92.40 \\
    3 & 98.27 & 81.31 & 97.05 \\
    4 & 98.30 & 90.53 & 98.95 \\
    5 & 98.35 & 93.92 & 99.75 \\
    10 & 98.35 & 97.42 & 100 \\

    \hline
  \end{tabular}
  \caption{Communicating private data through aggregated model in FL setting}
  \label{tab:fl_transfer}
\end{table}
\fi 

\subsection{Analyzing the Impact of Multiple Reads Under Idealized Assumptions}\label{optimizations}

\begin{figure}[ht]
    \centering 
    \includegraphics[width=2.1in]{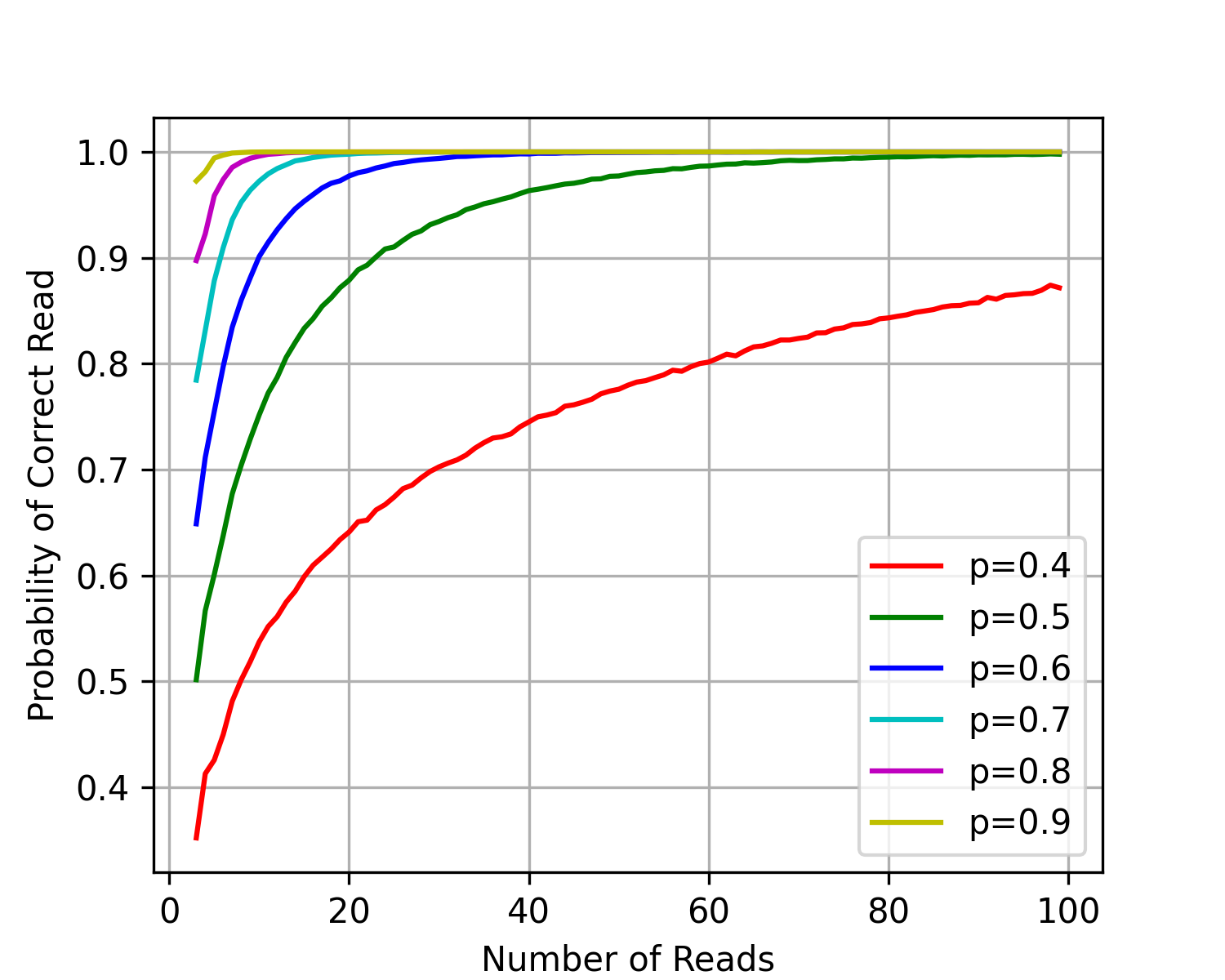} 
    \caption{Probability of correct value with multiple reads}
    \label{fig:multiple_read}
\end{figure}

  Briefly, we assume an underlying probability of the different classes such that the correct class has the highest probability; if this assumption is not true, then increasing the number of reads is not going to improve the probability of a correct read.  Assuming also that multiple reads represent independent trials, this becomes a multinomial experiment.  We model the multiple reads as a multinomial experiment derive using Monte Carlo simulation~\cite{mooney1997monte} estimates of the number of reads necessary to guarantee with a high probability that the most common class is the correct class. % using a bound derived from the Chernoff bound~\cite{chernoff1952measure}.  
The results are shown for different top-1 probabilities, and as we increase the number of reads in Figure~\ref{fig:multiple_read}.  We see that even with a few reads, the probability can be very high to get the correct output as the most commonly seen value.

\subsection{Optimized design considerations for Combinatorial Error Correction (CEC)}\label{cecdesign}
With every combination in CEC, there is a small chance of $\frac{1}{2^n}$ where $n$ is the number of CRC bits of aliasing, assuming a well-chosen CRC polynomial.  Aliasing is when a message  combination passes the CRC check, but is not the correct message.  The larger the size of the checksum block, the less the likelihood of aliasing.  However, larger CRC checks increase the overhead or, if amortized over more data blocks, increase the number of permutations needed before finding the correct message.  
%the more overhead we have to allow with the private data and more permutations (high computation complexity) that we have to check to reconstruct the private data. On the other hand, if we use small checksum block size, we end up with unlikely classes, and increase the possibility of aliasing (getting a wrong output that matches the checksum). The aliasing could occur by matching incorrect data cells with correct/wrong checksum cells or correct data cells with incorrect checksum cells (false positive); thus, 

Thus, we have to configure CEC to balance these considerations (i.e., computational complexity, accuracy/aliasing and overhead).  We evaluated both CRC8 and CRC12 for the error rates that we are encountering and found CRC12 to be more effective due to the significantly lower probability of aliasing.  %We found that CRC12 outperforms CRC8 in terms of accuracy with a little higher overhead as CRC12 has less probability of aliasing $(1/2^{12})$ whereas CRC8 has the probability of aliasing ($(1/2^{8})$ (detailed in appendix~\ref{CRC12vsCRC8}).
%According to our observations, both aliasing count and erroneous data block count rises more for the case of CRC8 in comparison to CRC12 while data block size increases or single cell top-1 accuracy decreases (detailed in appendix~\ref{CRC12vsCRC8}). 
 CRC16 or higher could provide even lower aliasing but come at higher storage and computational overhead.  %so on to get higher accuracy and low aliasing effect, but those could produce more overhead and more computational complexity while doing the permutation encoding. So, we choose CRC12 over others. %by considering a reasonable computational complexity and overhead along with a good accuracy while reconstructing the blocked private data in the receiver side.

%\noindent
%\textit{\textbf{Method for the computation of the data block size:}} 

A related challenge is how to size the data block given a chosen CRC algorithm.  %We refer to a combined block size as the sum of data and checksum block size. 
As the checksum block size is fixed in size, so ideally we would like to increase the size of the data to have a lower overall storage overhead.  However, the computational complexity rises with the number of included blocks as the number of permutations increases exponentially with the number of addresses in a block.  For example, if we choose the data block of size 8 and the checksum block of size 4 with top three (topK=3) most probable class for each cell, then we need to find  $3^{12}$ or over half a million permutations if we consider all possible permutations.  However, since we are limited in the number of permutations because of aliasing, we are able to consider only a small subset of the most probable permutations, resulting in lower correction success.
%(typically we consider only the most probable ones because the probability of aliasing increases with the number of permutations).
%(sorted based on the class probability of the cells) for a combined block of size 12. 
%As we have a confidence score for each of the 12 cells of the combined block, so we can assign a single numerical value for each  12 sized combined block by multiplying the confidence score of each of those 12 cells correspondingly. Finally, we set the priority of each permutation of the 12 sized combined block based on the corresponding numerical value; the permutation of the highest numerical value comes first for trying and so on. Assume, we would like to communicate D digits(3 bits each) over the channel and the combined block size is B. So, the total number of this combined block would be $ \lceil D/B \rceil $. 
%We observed that if we go for trying a large number of permutations for a combined block, it increases the possibility of aliasing. However, using our data extraction simulator, we observed that we could find most of the private data effectively within a certain depth of the permutation checking space based on the channel quality showed in table
Empirically, we find that the most efficient configurations based on the top-1 vary as shown in Table~\ref{tab:config}. 
\begin{table}[!tbh]
\begin{center}
\begin{tabular}{|c|c|c|c|} 
 \hline
 \textbf{Top 1 accuracy} & \textbf{Block size} & \textbf{Depth limit} & \textbf{topK}\\ 
 \hline
 $95<=x<=100$ & 7 & 350 & 3 \\ 
 \hline
 $90<=x<95$ & 5 & 450 & 4  \\
 \hline
 $x<90$ & 5 & 650 & 4 \\
 \hline
\end{tabular}
\caption{Configuration selection based on channel quality}
\label{tab:config}
\end{center}
\end{table}
\noindent

\subsection{Transferring Text and Random data through Lenet-5 and Alexnet using \ourapproach-C channel}\label{covert_experiment}

\begin{figure}[!tbh]
    \centering
    \includegraphics[width=\linewidth]{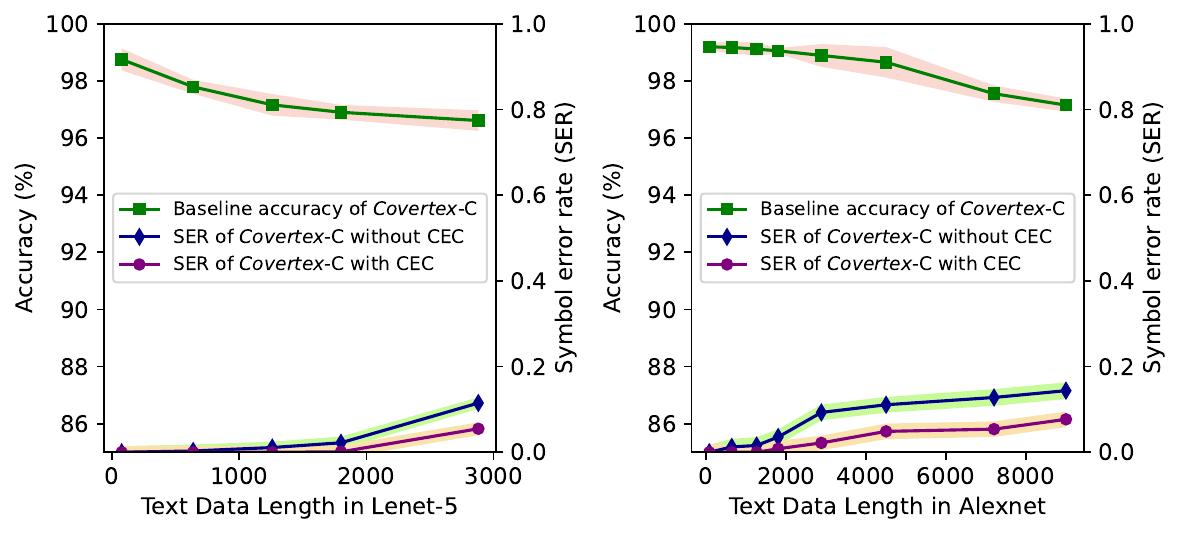} 
    \caption{With the increasing number of text data, baseline accuracy is degrading, and the symbol error rate is showing an upward trend as we approach the capacity.   }
    \label{text_lenet_5_alexnet_SER}
    %\nael{Mamun, can you please remove the red line?  If you want to keep it, please remove the dots (triangles) on it since there is no experimental value for each data length.  The line value is already implied from left most value.  Please do this for all these figures (15, 16, 17). Can you please also use CEC instead of ECC? Thanks.}
%\nael{Also, please use the word "accuracy" in the baseline curves so that we know to look at the left x-axis for them.}\mamun{I updated the figure..}
\end{figure}

Figure~\ref{text_lenet_5_alexnet_SER} shows that we can send up to 2880 addresses (3 bit digit each) with baseline accuracy degradation of 2.16\% (from 98.74 to 96.61) with Lenet-5 trained with MNIST. For the Alexnet trained with MNIST, We could able to send up to 9000 digits with baseline accuracy degradation of 2.05\% ( from 99.18 to 97.15) shown in Figure~\ref{text_lenet_5_alexnet_SER}. For both models, we notice that baseline accuracy degrades with the increasing size of the private text data length, as we near the capacity of the model. %The more the textdata increases, gradually we push to the capacity of the both model. We also empirically found that for the models with a higher number of parameters, we can send the same number of text data with low degradation of baseline accuracy (later we will show this for Resnet50  also). 

%Moreover, to get the advantage of combinatorial error correction (CEC) we added the checksum with the text data from the sender side and Figure~\ref{text_lenet_5_alexnet_SER} clearly shows that we achieved better stored data accuracy/lower symbol error rate for transferring text data using CEC. 

%Text data has built-in redundancy since ASCII values are concentrated in a range that the network can efficiently learn~\cite{mei2005discovering,delgado2002mining}. 

\begin{figure}[!tbh]
    \centering
    \includegraphics[width=\linewidth]{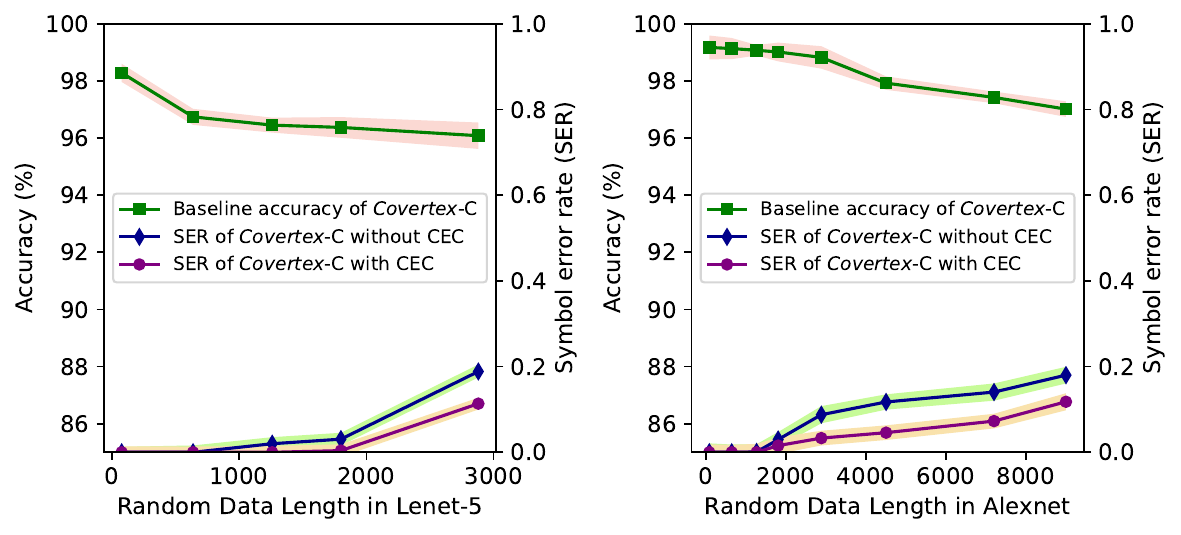} 
    \caption{Random data: Baseline accuracy degrades, and symbol error rate increases as we approach capacity.} %\textcolor{blue}{it might be a plus to illustrate with a text sent through the model (even in the Appendix if we don't have space)}\mamun{We are really out of space i guess. We are planning to omit the appendix and bring important figures from appendix to the paper.}}
    \label{random_lenet_5}
\end{figure}

To get a true measure of capacity, we also communicated random data to the receiver using \ourapproach-C. % Because there might be some frequent letter which may help the model to learn the pattern. We transferred random data through (1) Lenet-5, and (ii) Alexnet both trained with MNIST dataset. 
We used different lengths of random data and report the symbol error rate of the channel as well as the accuracy degradation of baseline for both models shown in Figure~\ref{random_lenet_5}.

 %Note that we added the ECC with the random data from the sender side to show the advantage of our combinatiorial error correction technique in a covert channel for achieving the data in better accuracy than a covert ML channel without any ECC. In the experimental results showed in  figure~\ref{random_lenet_5}, we used same ECC configurations that we used for text data before for Lenet-5 and modified Alexnet correspondingly. 

 Figure~\ref{random_lenet_5} shows the results for sending random data, for message sizes similar to the text experiment.  The behavior shows similar overall patterns.  The accuracy drop for the same size was marginally higher (e.g., up to 2.66\% drop from 98.74 to 96.08 on Lenet-5) as we increase the size of the random message, since the entropy of the random message is higher than that of text.  %The message size is equal to the sizes we used for the text experiments.  For Alexnet, We sent up to 9000 3-bit digits of random data with baseline accuracy degradation of 2.17\% ( from 99.18 to 97.01). For both models, baseline accuracy degrades gradually with the increasing size of the random data length.
%By increasing the length of random data, gradually we are pushing to the capacity of the model and we can see the degradation of the accuracy on random data while we increase the length of the private data.  
%We also observed that the random data extraction accuracy degrades a little faster compared to text data. 
As with text, CEC improves the symbol error rate. 
 %However, our error correcting code (ECC) makes the degradation slow, thereby minimizing the symbol error rate of the covert ML channel.

%Like the text data, here also we trained the model for 200 training epoch in the sender side to communicate the corresponding length of random data with the receiver showed in figure~\ref{random_lenet_5}. 

\end{document}